\documentclass{article} 
\usepackage{iclr2023_conference,times}

\usepackage{amsmath,amsfonts,bm}

\def\eqref#1{equation~\ref{#1}}

\def\1{\bm{1}}

\DeclareMathAlphabet{\mathsfit}{\encodingdefault}{\sfdefault}{m}{sl}
\SetMathAlphabet{\mathsfit}{bold}{\encodingdefault}{\sfdefault}{bx}{n}

\usepackage[utf8]{inputenc} 
\usepackage[T1]{fontenc}    
\usepackage{url}            
\usepackage{booktabs}       
\usepackage{amsfonts}       
\usepackage{nicefrac}       
\usepackage{microtype}      
\usepackage{xcolor}         
\usepackage{subfig}
\usepackage{graphicx}
\usepackage{amsmath}         
\usepackage{csquotes}
\usepackage{tikz} 
\usepackage[ruled,vlined]{algorithm2e}
\tikzstyle{block} = [draw, fill=white!40, rectangle, minimum height=2.5em, minimum width=4em]
\tikzstyle{sum} = [draw, fill=white, circle, scale=0.005, node distance=0.5cm]
\tikzstyle{input} = [coordinate]
\tikzstyle{output} = [coordinate]  
\tikzstyle{pinstyle} = [pin edge={to-,thin,black}]
\usetikzlibrary{shapes,arrows}
\usepackage{adjustbox}
\usepackage{multirow}
\usepackage{bbm}

\iclrfinalcopy

\title{Vision HgNN: An Electron-Micrograph is Worth Hypergraph of Hypernodes}

\author{
  Sakhinana Sagar Srinivas$^{1}$\thanks{Conceived, designed, implemented the research(programmed the software) and drafted the manuscript}, \hspace{1mm}[Sreeja Gangasani$^{2}$, Rajat Kumar Sarkar$^{1}$]\thanks{Performed computational experiments, interpretation, and visualization analysis of the results},\\\textbf{Venkataramana Runkana$^{1}$}\\
  TCS Research$^{1}$, Indian Institute of Technology$^{2}$ \\
  \texttt{sagar.sakhinana@tcs.com, rajat.sarkar1@tcs.com} \\ \texttt{111901023@smail.iitpkd.ac.in}, 
  \texttt{venkat.runkana@tcs.com}
}

\begin{document}

\maketitle

\vspace{-2mm}
\begin{abstract} 
\vspace{-2mm}
Material characterization using electron micrographs is a crucial but challenging task with applications in various fields, such as semiconductors, quantum materials, batteries, etc. The challenges in categorizing electron micrographs include but are not limited to the complexity of patterns, high level of detail, and imbalanced data distribution(long-tail distribution). Existing methods have difficulty in modeling the complex relational structure in electron micrographs, hindering their ability to effectively capture the complex relationships between different spatial regions of micrographs. We propose a hypergraph neural network(HgNN) backbone architecture, a conceptually alternative approach, to better model the complex relationships in electron micrographs and improve material characterization accuracy. By utilizing cost-effective GPU hardware, our proposed framework outperforms popular baselines. The results of the ablation studies demonstrate that the proposed framework is effective in achieving state-of-the-art performance on benchmark datasets and efficient in terms of computational and memory requirements for handling large-scale electron micrograph-based datasets.
\end{abstract}

\vspace{-7mm}
\section{Introduction} 
\begin{figure}[htbp]
     \centering
     \subfloat[High intra-class dissimilarity in electron micrographs of \textit{MEMS} device.]{\includegraphics[width=0.18\textwidth]{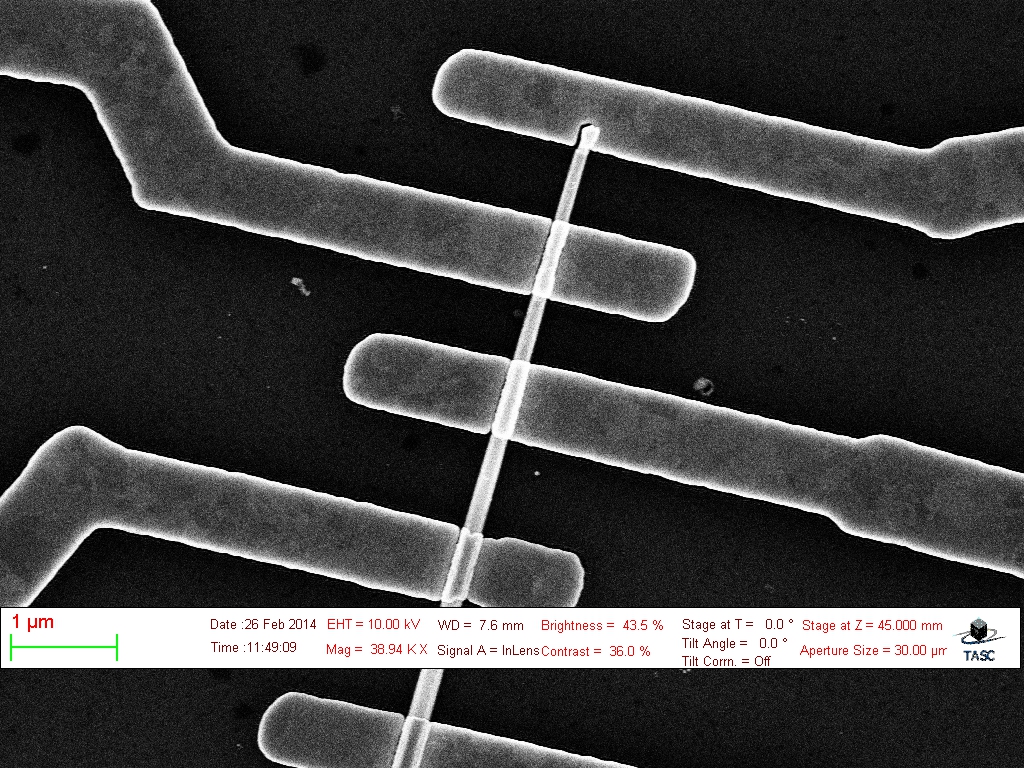}
     \includegraphics[width=0.18\textwidth]{Images/variance/h.jpg}
     \includegraphics[width=0.18\textwidth]{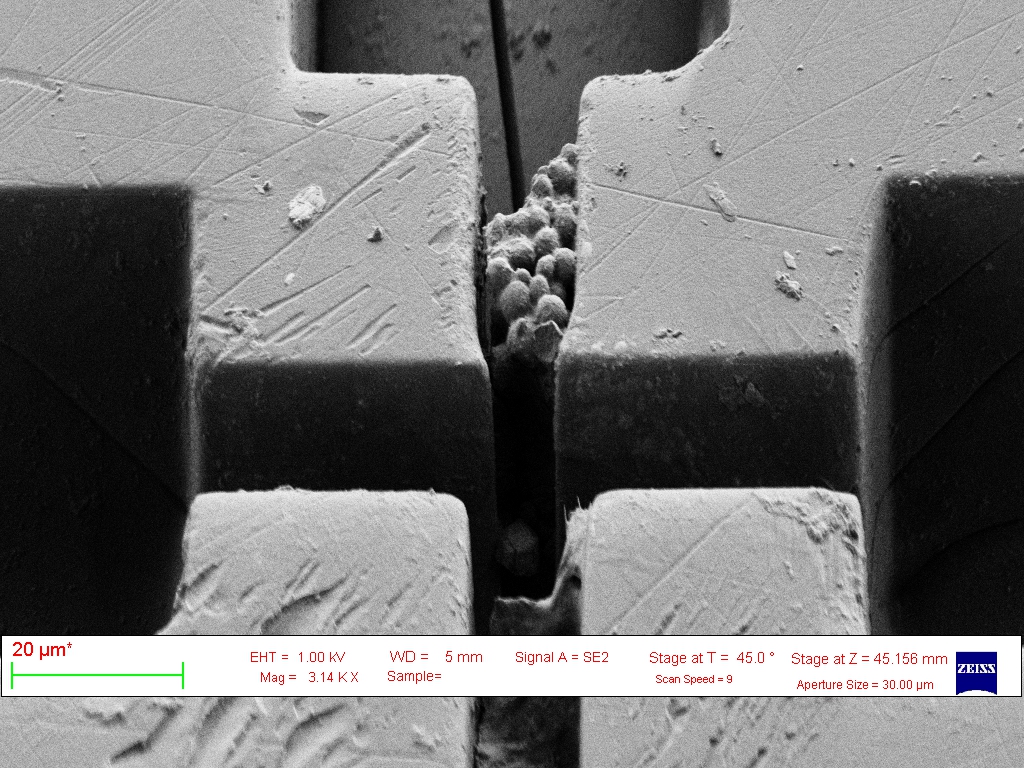}
     \includegraphics[width=0.18\textwidth]{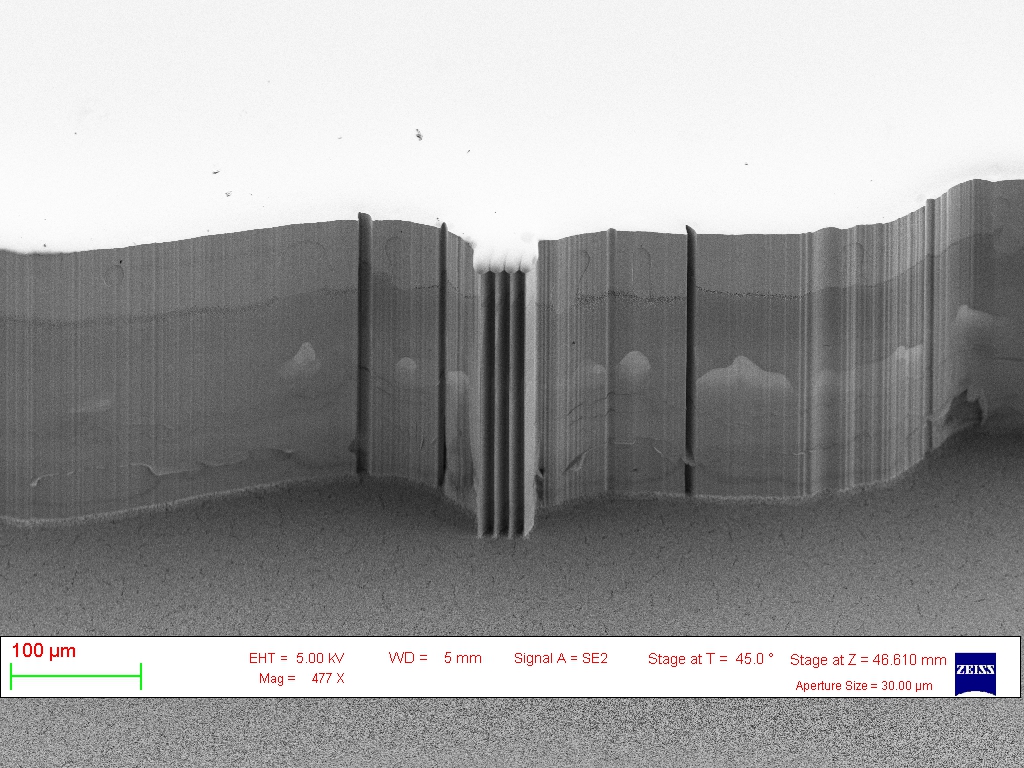}
     }
     \vspace{0mm}
     \qquad
     \subfloat[The high inter-class similarity in different material categories(left to right: \textit{films, powder, particles, porous sponges}).]{\includegraphics[width=0.18\textwidth]{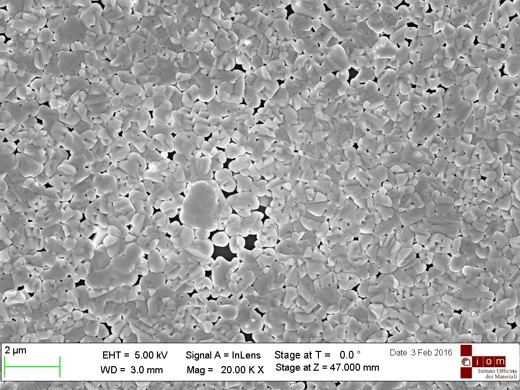}
     \includegraphics[width=0.18\textwidth]{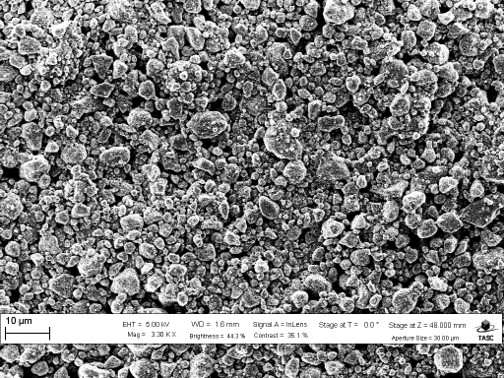}
     \includegraphics[width=0.18\textwidth]{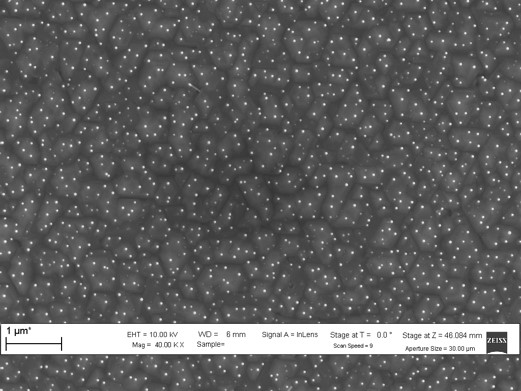}
     \includegraphics[width=0.18\textwidth]{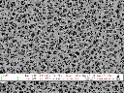}
     }
     \vspace{0mm}
     \qquad
     \subfloat[Multi-spatial scales of patterns in electron micrographs of \textit{particles}.]{\includegraphics[width=0.18\textwidth]{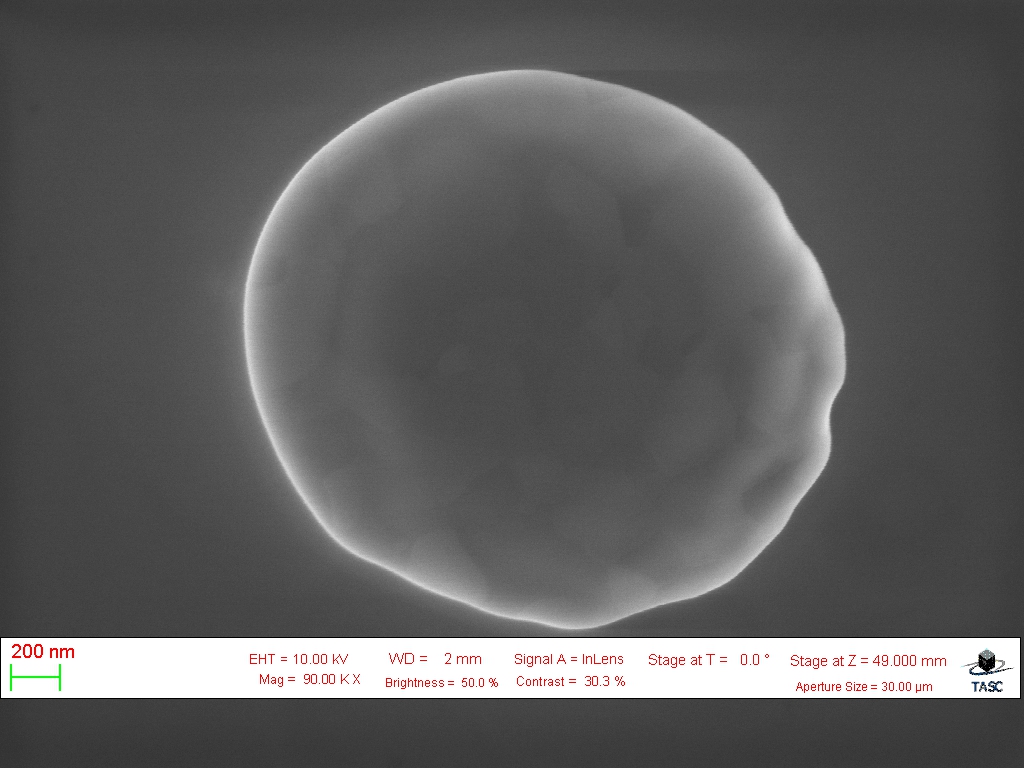}
     \includegraphics[width=0.18\textwidth]{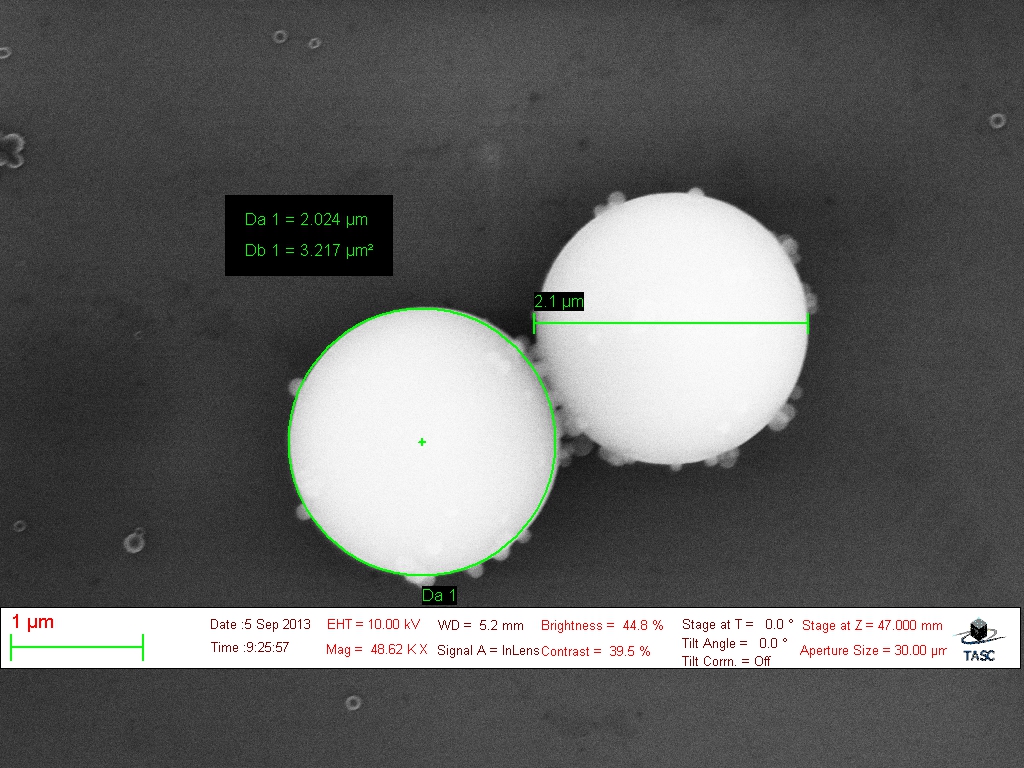}
     \includegraphics[width=0.18\textwidth]{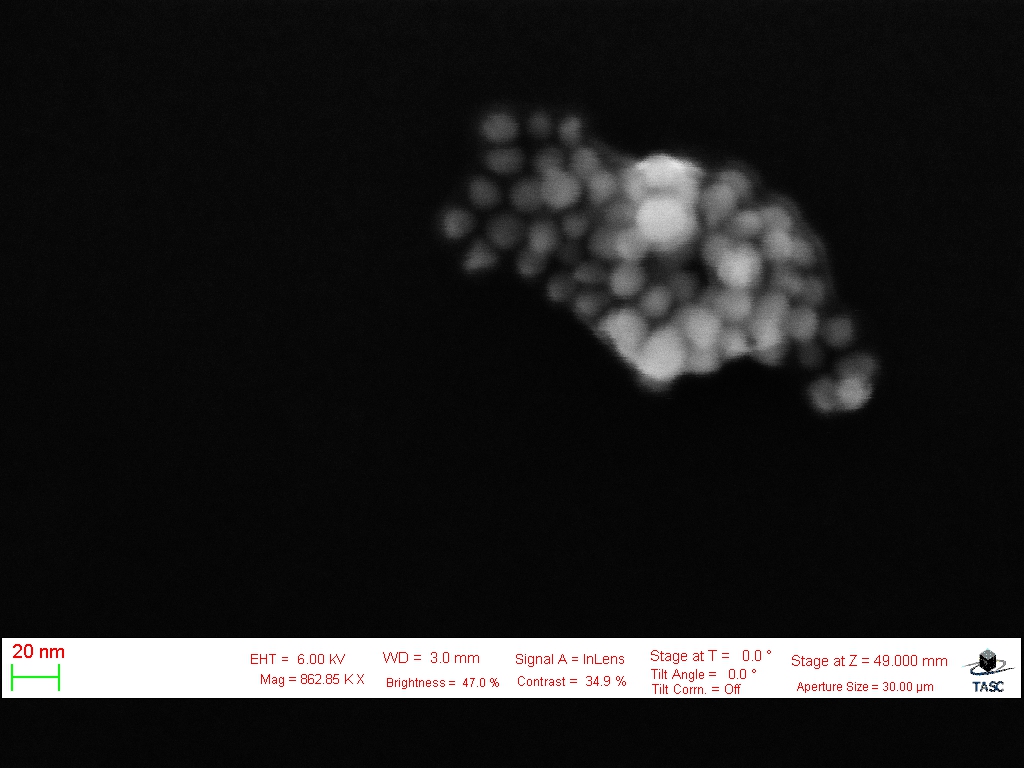}
     \includegraphics[width=0.18\textwidth]{Images/patterns/b.jpg}}
     \vspace{-2mm}
     \caption{The figure depicts the various challenges in the electron micrograph classification task on the SEM dataset(\cite{aversa2018first}).}
     \label{fig:challenges}
     \vspace{-2mm}
\end{figure}

\vspace{-5mm}
Accurate development, characterization, and testing of miniaturized semiconductor devices are essential in leading-edge chip design to ensure their proper functioning and performance. State-of-the-art imaging and analysis techniques(\cite{holt2013sem}) play a critical role in the fabrication, inspection, and testing of the next-generation miniaturized semiconductor devices, such as those with a feature size of 7nm or smaller, as they help ensure their quality and reliability. The advanced imaging of miniature devices in the semiconductor industry typically utilizes a broad spectrum of electron beam tools, including Scanning Electron Microscopy(SEM), Transmission Electron Microscopy(TEM), Reflective Electron Microscopy(REM), and others. Electron microscopes, as ultra-modern imaging tools, produce high-magnification and high-resolution images of material specimens, known as electron micrographs, to perform microstructural characterization or identification of materials, which is essential for the accurate design fabrication, and evaluation of miniaturized semiconductor devices. However, classifying electron micrographs is challenging due to high intra-class variance, low inter-class dissimilarity, and multiple spatial scales of visual patterns. Figure \ref{fig:challenges} illustrates the various challenges in the automatic nanomaterial identification task. The de facto standard neural-network architectures for vision tasks such as ConvNets(\cite{iandola2016squeezenet}, \cite{he2016deep}), Vision transformers(ViTs, \cite{dosovitskiy2020image}, \cite{SwinT}, \cite{ConViT}, \cite{Crossvit}), hybrid architectures(\cite{CVT}, \cite{Levit}, \cite{PVT}) and MLP-based vision models(\cite{tolstikhin2021mlp}, \cite{touvron2021resmlp}) do not explicitly model the higher-order dependencies between the multiple grid-like patches(also referred to as tokens) of the electron micrographs. Nevertheless, this work aims to explore an alternative effective, efficient neural-network architecture beyond traditional methods for modeling the fine-grained interrelations among the spatially and semantically dependent regions(patches) of the electron micrographs for automatic nanomaterial identification tasks via the hypergraph framework. We utilize the hypergraphs as a mathematical model(\cite{ouvrard2020hypergraphs, feng2019hypergraph, gao2022hgnn, yadati2019hypergcn}) for a  structured representation of the electron micrographs to learn the hierarchical relations among the spatial regions(patches) unconstrained by their spatial location in the micrograph.  We begin with a hypothesis, that the electron micrographs have an inherent hypergraph structure that capture a wide range of structural and property information from sub-hypergraphs of various sizes, and represents the higher-order dependencies between the patches in the electron micrographs. We present the Vision Hypergraph Neural Networks(for short, Vision-HgNN) designed to automatically encode the dominant structural and feature information of the visual hypergraphs and then learn relation structure-aware hypergraph-level embeddings to perform effective neural relational inference on the downstream multi-class classification task. The proposed visual hypergraph representation learning framework is designed to identify discrete visual elements(low-level entities) and their higher-order dependencies, and to prune redundant visual elements to learn an efficient representation of scale-variant visual elements(high-level entities) for improved perception and reasoning of the visual content in hypergraph-structured micrographs to enhance classification performance. The proposed framework is intended to offer better generalization and scalability for large-scale electron microscopy image corpus-based classification tasks.

\vspace{-3mm}

\vspace{-5mm} 
\section{Problem statement} 
\vspace{-3mm}
Consider a dataset consisting of visual hypergraph-label pairs $(\mathcal{G}_{i}, y_{i})$ where the ground-truth label of $\mathcal{G}_{i}$ is denoted by $y_{i}$.
The objective of the classification task is to learn a novel mapping neural network function $f: \mathcal{G}_{i} \rightarrow y_{i}$ that maps the discrete visual hypergraphs to the set of predefined categories.

\vspace{-4mm}
\begin{figure}[htbp]
    \centering
    \subfloat[Grid]{{\includegraphics[scale=0.17]{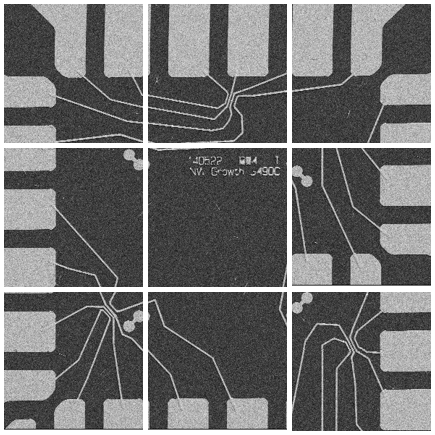}}}
    \hspace{-0.5mm}
    \vspace{2mm}
    \subfloat[Sequence]{{\includegraphics[scale=0.12]{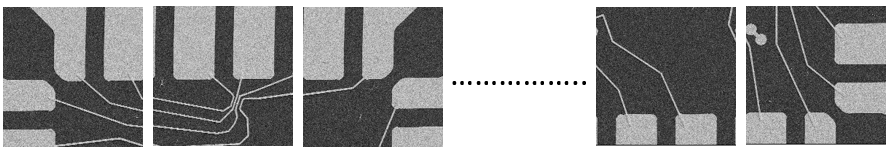}}}
    \hspace{-0.5mm}
    \subfloat[Graph]{{\includegraphics[scale=0.17]{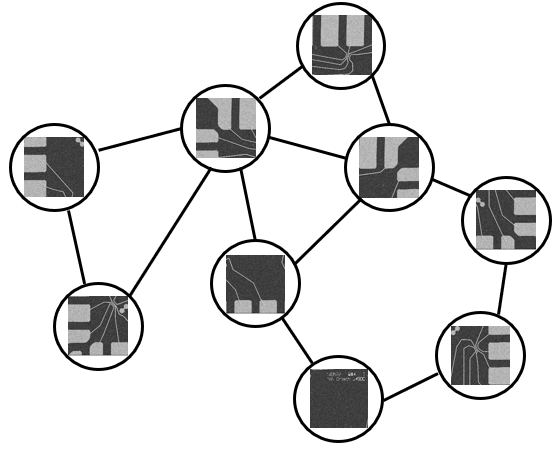}}}
    \hspace{-0.5mm}
    \subfloat[Hypergraph]{{\includegraphics[scale=0.17]{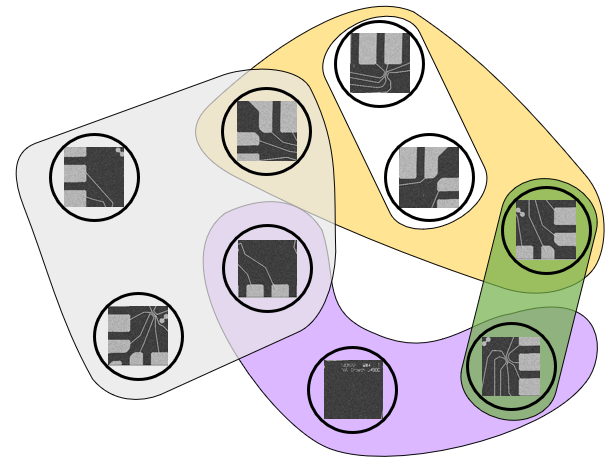}}}
    \vspace{-4.7mm} 
    \caption{For illustration purpose, an electron micrograph(MEMS device, \cite{aversa2018first}) was split into $\text{3} \times \text{3}$ patches. This figure depicts a regular grid, a sequence, a graph, and a hypergraph representation of an electron micrograph. (a) ConvNets operate on a grid of pixels, (b) ViTs operate on a sequence of grid-like patches, (c) GNNs operate on visual graphs where patches are viewed as nodes, and (d) HgNNs operate on visual hypergraphs where the patches represent the hypernodes to perform classification tasks. The visual graph and hypergraph structure representations are learned through the nearest neighbor search algorithm. They are linked based on the visual content and are not necessarily determined by their spatial location in the micrograph. The edges in the graph model pair-wise relations among the patches, while hyperedges model multi-dyadic relationships.}
    \label{fig:a}
\end{figure}

\clearpage
\newpage

\vspace{-4mm} 
\section{Proposed Approach} 
\label{pralgo}
\begin{figure}[htbp]
\centering
\begin{tikzpicture}[auto, node distance=2cm,>=latex']
    \node [input, name=input] {};
    \node [sum, right of=input] (sum) {};
    \node [block, thick, right of=sum] (controller) {$\text{HgSL}$};
    \node [block, thick, right of=controller, node distance=2cm] (system) {$\text{HgAT}$};
    \node [block, thick, right of=system, node distance=2cm] (system_) {$\text{HgT}$};
    \node [block, thick, right of=system_, node distance=2cm] (system__) {$\text{HgRo}$};
    
    \draw [->] (sum) -- node[label={[xshift=-1.25cm, yshift=-0.25cm] $\text{Electron micrograph,} \hspace{1mm} i$}] {} (controller); 
    \draw [->] (controller) -- node {$\mathcal{G}_{i}$} (system); 
    \node [output, right of=system__] (output) {};
     
    \draw [->] (controller) -- node[label={[xshift=0.0cm, yshift=-0.2cm]$ $}] {} (system);
    \draw [->] (system) -- node[label={[xshift=0.0cm, yshift=-0.2cm]$ $}] {} (system_);
    \draw [->] (system_) -- node[label={[xshift=0cm, yshift=-0.2cm]$ $}] {} (system__);
    \draw [->] (system__) -- node[label={[xshift=0cm, yshift=-0.2cm]$ $}] {$y^{p}_{i}$} (output);
    
\end{tikzpicture}
\vspace{-2mm}
\caption{The isotropic $\text{Vision-HgNN}$ architecture. $y^{p}_{i}$ denotes the model predictions.} \label{fig:b}
\end{figure}
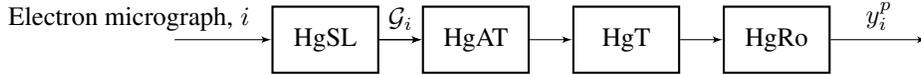

\vspace{-4mm}
As illustrated in Figure \ref{fig:b}, our framework consists of the following modules. (a) hypergraph structure learning module, for brevity, \textbf{HgSL} backbone, to learn the discrete visual hypergraph representations of the electron micrographs through pairwise proximity function. (b) a local and global neighborhood connectivity-driven hypergraph attention network, for brevity, the \textbf{HgAT} backbone is designed to capture short, and moderate-range dependencies, i.e., encapsulates the hypergraph's structural and feature information in the hypernode-level embeddings. (c) a self-attention mechanism-based-hypergraph transformer network, for brevity, an \textbf{HgT} backbone with no hypergraph spatial priors to learn all pairwise hypernode interactions for better learning of the long-range pairwise dependencies. (d) the hypergraph read-out module, for brevity, the \textbf{HgRo} backbone, performs the global average pooling that collapses hypernode-level embeddings to obtain the single hypergraph-level embedding. (e) a linear projection layer and a normalized exponential function transform the hypergraph-level embedding to a multinomial probability distribution over the predefined electron micrograph categories to predict the visual hypergraph category.

\vspace{-3mm}

\vspace{-5.3mm}
\subsection{Hypergraph Structure Learning(HgSL)}
\vspace{-2mm}
The HgSL operates in two phases. At first, it performs tokenization of electron micrographs. Next, it optimizes the discrete visual hypergraph structure through a differentiable approach to learn the more robust and optimal representation through the nearest neighbor search technique and formulate the posterior classification task as message-passing schemes with hypergraph neural networks. 

\vspace{-3mm}
\subsubsection{Electron Micrograph Tokenizer}
\vspace{-3mm}
We split an electron micrograph with the size $h \times w \times c$, where $(h,w)$ is the resolution of the RGB image, and $c$ is the number of channels into non-overlapping $n$ uniform patches, where the size of each patch is $p \times p \times c$ and $p$ is patch size. We reshape the patches to obtain feature matrix \resizebox{.12\textwidth}{!}{$\mathbf{X}^{\prime} \in \mathbb{R}^{n \times p^{2}c}$}. We linearly transform the feature matrix, \resizebox{.025\textwidth}{!}{$\mathbf{X}^{\prime}$}, through a trainable embedding layer $\mathbf{E}$ to compute a refined feature matrix \resizebox{.105\textwidth}{!}{$\mathbf{X} \in \mathbb{R}^{n \times d}$} as described below,

\vspace{-2mm}
\resizebox{0.925\linewidth}{!}{
\begin{minipage}{\linewidth}
\begin{equation}
\mathbf{X} = \mathbf{X}^{\prime}\mathbf{E}; \hspace{1mm}  \mathbf{E} \in \mathbb{R}^{p^{2}c \times d}\hspace{1mm} \label{eq:hsl1}
\end{equation}
\end{minipage}
}

\vspace{-1mm}
The row $i$ of the feature matrix \resizebox{.17\textwidth}{!}{$\mathbf{X} = \left[\mathbf{x}_{p}^1; \ldots; \mathbf{x}_{p}^n\right]$} represents the (low-dimensional) feature representation for patch  \resizebox{.23\textwidth}{!}{$\boldsymbol{x}^{i}_{p} \in \mathbb{R}^{d}$, $i = 1, 2, \ldots,n$}, where $d$ is the predefined feature dimension.

\vspace{-4mm}
\subsubsection{Hypergraph Representation}
\vspace{-3mm}
We represent the patches as the unordered hypernodes of an undirected visual hypergraph denoted as \resizebox{.195\textwidth}{!}{$\mathcal{V}=\{v_{1},v_{2},\ldots,v_{n}\}$}. For each hypernode $v_{i}$, we form an undirected hyperedge $e_{p}$ from the hypernode $v_{i}$ to $v_{j}$; if $v_{j}$ is among the top-K visual-semantic-nearest neighbors of $v_{i}$. Thus we obtain $n$ hyperedges incident with $\text{K}+1$ non-repeating hypernodes. The hyperedges describe the relations and capture more complex relationships and interdependencies among hypernodes in the visual hypergraphs. We then obtain a hypergraph \resizebox{.13\textwidth}{!}{$\mathcal{G}= (\mathcal{V}, \mathcal{E}, \mathbf{X})$} where \resizebox{.195\textwidth}{!}{$\mathcal{E}=\{e_{1},e_{2},\ldots,e_{n}\}$} denotes the set of hyperedges and \resizebox{.10\textwidth}{!}{$\mathbf{X} \in \mathbb{R}^{n \times d}$} is the hypernode feature matrix. Each row $i$ in \resizebox{.02\textwidth}{!}{$\mathbf{X}$} represents the hypernode feature vector, $x_{v_{i}} \in \mathbb{R}^{d}$. Note: $x_{v_{i}}$ is the patch feature representation, $x^{i}_{p}$. The incidence matrix, \resizebox{.11\textwidth}{!}{$\mathbf{H} \in \mathbb{R}^{n \times n}$}, describes the hypergraph structure. \resizebox{.09\textwidth}{!}{$\mathbf{H}_{i, \hspace{0.5mm}p}=1$} if the hyperedge $p$ incident with hypernode $i$ and otherwise 0.  The hyperparameter $\text{K} < n$ determines the sparsity of the visual hypergraph. Let \resizebox{.215\textwidth}{!}{$\mathcal{N}_{p, i}=\big\{v_{i} | \mathbf{H}_{i, p} = 1\big\}$} represent the subset of hypernodes $v_{i}$ incident with any hyperedge $p$. The intra-edge neighborhood of the hypernode $i$ is given by \resizebox{.065\textwidth}{!}{$\mathcal{N}_{p,i} \backslash i$}. It is a localized group of perceptually similar patches and captures higher-order relationships. The inter-edge neighborhood of hypernode $i$, \resizebox{.215\textwidth}{!}{$\mathcal{N}_{i, p} = \big\{e_{p}|\mathbf{H}_{i, p} = 1\big\}$}, spans the spectrum of the set of hyperedges $e_{p}$ incident with hypernode $i$. There is no natural ordering of the hypernodes in the hypergraph. To preserve patch-locality information in the main hypergraph, we linearly add the trainable position embeddings($\mathbf{E}_{pos}$) to the hypernode feature vectors to enable position awareness. The HgSL module computes the hypernode's positional embeddings based on the intra- and inter-edge neighborhood in the hypergraph. Equation \ref{eq:hsl} shows the transformed feature vector of the hypernodes.

\vspace{-2mm}
\resizebox{1\linewidth}{!}{
\begin{minipage}{\linewidth}
\begin{equation}
\left[\mathbf{x}_{v_{1}}; \ldots; \mathbf{x}_{v_{n}}\right] =\left[\mathbf{x}_{v_{1}}, \ldots, \mathbf{x}_{v_{n}}\right]\hspace{-0.5mm} + \mathbf{E}_{p o s};\hspace{1mm}  \mathbf{E}_{pos} \in \mathbb{R}^{n \times d}  \label{eq:hsl}
\end{equation}
\end{minipage}
}

\subsection{Model Architecture}
\begin{figure}[htbp]
\hspace*{-28mm}    
    \includegraphics[width=1.425\textwidth]{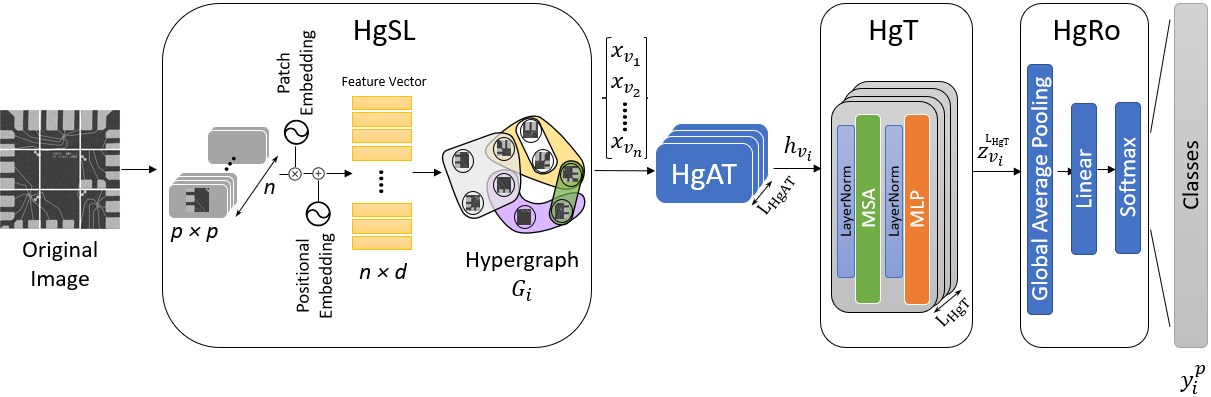}
    \vspace{-12mm}
    \caption{For illustration, we split the electron micrograph into $3 \times 3$ patches. We represent the electron micrograph as a patch-attributed visual hypergraph. The framework presents an end-to-end visual hypergraph representation learning with Hypergraph Neural Networks for categorization tasks.}
    \label{fig:imagetok}
\end{figure}

\vspace{-3mm}
Figure \ref{fig:imagetok} depicts the Vision-HgNN framework. The Hypergraph attention network(HgAT) performs the message-passing schemes on the visual hypergraphs  to obtain (low-dimensional) hypernode embeddings. The Hypergraph transformer(HgT)  utilizes the self-attention mechanism for transforming the hypernode embeddings determined by the HgAT operator to compute refined hypernode embeddings(\resizebox{.135\textwidth}{!}{$\mathbf{z}^{L_{\text{HgT}}}_{v_{1}}, \dots, \mathbf{z}^{L_{\text{HgT}}}_{v_{n}}$}). We discussed the HgAT and HgT operators in the appendix. The HgRo module performs the average pooling of hypernode embeddings(\resizebox{.135\textwidth}{!}{$\mathbf{z}^{L_{\text{HgT}}}_{v_{1}}, \dots, \mathbf{z}^{L_{\text{HgT}}}_{v_{n}}$}) to obtain hypergraph-level embedding $\mathbf{z}^{L_{\text{HgT}}} \in \mathbb{R}^{d}$. We apply a linear projection and softmax to transform $\mathbf{z}^{L_{\text{HgT}}}$ for determining the model predictions \resizebox{.25\textwidth}{!}{$y^{p}_{i} =\text{softmax}\big(\text{W}^{out}\mathbf{z}^{L_{\text{HgT}}}\big)$}, where \resizebox{.125\textwidth}{!}{$\text{W}^{out} \in \mathbb{R}^{d \times d}$}.

\vspace{-4mm}
\subsection{Algorithmic Architecture}
\vspace{-3mm}

\vspace{-7mm} 
\section{Experiments and Results}\label{sec:expresults}
\vspace{-3mm}
\subsection{Datasets}
\vspace{-3mm}
We conduct experiments on the SEM dataset(\cite{aversa2018first}) for automatic nanomaterials identification. The human-annotated dataset contains a set of 10 categories belonging to a wide range of nanomaterials spanning a broad range of particles, nanowires, patterned surfaces, etc., for a total of $\approx$21,283 electron micrographs. The initial experimental results are reported by \cite{modarres2017neural} on the subset of the complete dataset. Due to the unavailability of the subset dataset publicly, we conducted experiments on the original dataset(\cite{aversa2018first}), which contains 12$\%$ more samples. The dataset curators(\cite{aversa2018first}) had not shared the predefined train/validation/test splits, so we leveraged the k-fold cross-validation technique to evaluate our model performance for competitive benchmarking with the varied baseline models. Additionally, we utilized several open-source material benchmark datasets to demonstrate the effectiveness of our proposed method.

\vspace{-3mm}
\subsection{Benchmarking algorithms} %
\vspace{-2mm}
We train the $\text{Vision-HgNN}$ framework through a supervised learning approach for joint visual hypergraph inference and category prediction of micrographs. Table \ref{tab:cstudy} provides a performance comparison of the Vision-HgNN framework with other baseline models, which include ConvNets, GNNs(\cite{rozemberczki2021pytorch, Fey/Lenssen/2019}), and ViTs(\cite{philvformer, neelayvformer}) architectures. In addition, we utilize the different self-supervised learning algorithms: Vision Contrastive Learning(VCL, \cite{susmelj2020lightly})) and Graph Contrastive Learning(GCL, \cite{Zhu:2021tu})) for comparison with our proposed method. We ensure a fair and rigorous comparison between the Vision-HgNN framework and the baseline algorithms by generating the results under identical experimental settings. The evaluation metric is the  Top-N accuracy, where $N \in \{1, 2, 3, 5\}$. The standard deviation values are less than at most 4$\%$ of the mean value. The proposed method demonstrates the best

\vspace{-4mm}
\begin{table}[htbp]
\centering
\renewcommand{\arraystretch}{1.0}
\setlength{\tabcolsep}{2.5pt}
\resizebox{0.75\textwidth}{!}{%
\subfloat{%
\begin{tabular}{cc|cccccc}
\hline
\multicolumn{2}{c|}{\textbf{Algorithms}}                      &\textbf{Parameters}                & \textbf{Top-1} & \textbf{Top-2} & \textbf{Top-3} & \textbf{Top-5}  \\ \hline
\multicolumn{1}{c|}{\multirow{6}{*}{\rotatebox[origin=c]{90}{\textbf{ConvNets}}}} & AlexNet(\cite{krizhevsky2017imagenet})     & 57.0M & 0.519	& 0.593	& 0.681	& 0.785             &                     \\
\multicolumn{1}{c|}{}                                          & DenseNet(\cite{huang2017densely})     & 0.24M  & 0.521	& 0.722	& 0.869	& 0.912
             &             \\
\multicolumn{1}{c|}{}                                          & ResNet(\cite{he2016deep})       & 0.27M  & 0.524	& 0.773	& 0.904	& 0.915 &             \\
\multicolumn{1}{c|}{}                                          & VGG(\cite{simonyan2014very})          & 34.4M  &  0.523	& 0.657	& 0.729	& 0.783 &                     \\
\multicolumn{1}{c|}{}                                          & GoogleNet(\cite{szegedy2015going})    & 0.26M  & 0.573	& 0.857	& 0.914	& 0.942
              \\
\multicolumn{1}{c|}{}                                          & SqueezeNet(\cite{iandola2016squeezenet})   & 0.74M  & 0.467	& 0.476	& 0.621	& 0.684
              \\ \hline
\multicolumn{1}{c|}{\multirow{12}{*}{\rotatebox[origin=c]{90}{\textbf{Vision Transformers(ViTs)}}}}        & CCT(\cite{hassani2021escaping})          & 0.41M    & 0.586	& 0.795	& 0.886	& 0.952           &             \\
\multicolumn{1}{c|}{}                                          & ConViT(\cite{ConViT})       & 0.60M    & 0.596	& 0.724	& 0.836	& 0.941                         \\
\multicolumn{1}{c|}{}                                          & PVTC(\cite{PVT})         & 1.30M    & 0.572	& 0.758	& 0.826	& 0.917
             &            \\
\multicolumn{1}{c|}{}                                          & SwinT(\cite{SwinT})       & 27.8M    & 0.658	& 0.763	& 0.904	& 0.928  \\
\multicolumn{1}{c|}{}                                          & VanillaViT(\cite{dosovitskiy2020image})   & 1.79M    & 0.638	 & 0.834	& 0.868	 & 0.943              \\
\multicolumn{1}{c|}{}                                          & CaiT(\cite{CaiT})         & 0.38M    & 0.627	 & 0.741  & 0.879	& 0.945             &            \\
\multicolumn{1}{c|}{}                                          & LeViT(\cite{Levit})        & 16.8M   & 0.625	 & 0.776	& 0.863	 & 0.957  \\
\multicolumn{1}{c|}{}                                          & NesT(\cite{Nest})         & 16.1M      & 0.643	& 0.835	& 0.923	& 0.952 &             \\
\multicolumn{1}{c|}{}                                          & PatchMerger(\cite{PatchMerger})  & 3.26M    & 0.561	& 0.703	& 0.842	& 0.939 \\
\multicolumn{1}{c|}{}                                          & RegionViT(\cite{Regionvit})    & 12.2M    & 0.589	& 0.812	& 0.863	& 0.936                         \\
\multicolumn{1}{c|}{}                                          & T2TViT(\cite{T2TViT})      & 10.3M    & 0.672	& 0.841	& 0.911	& 0.927 \\
\multicolumn{1}{c|}{}                                          & ViT-SD(\cite{ViT-SD})          & 4.47M     & 0.609	& 0.746	 & 0.873	& 0.949                         \\ 
\hline
\multicolumn{1}{c|}{}                                          & \textbf{Vision-HgNN}     &  0.91M    &    \textbf{0.819}	& \textbf{0.864}	& \textbf{0.942}	& \textbf{0.994}               &                     \\ \hline
\end{tabular}}
}
\vspace{1mm}
\end{table}

\vspace{-4mm}
\begin{table}[htbp]
\centering
\renewcommand{\arraystretch}{1.0}
\setlength{\tabcolsep}{2.5pt}
\resizebox{0.8\textwidth}{!}{%
\qquad
\subfloat{%
\begin{tabular}{cc|cccccc}
\hline
\multicolumn{2}{c|}{\textbf{Algorithms}}                           &\textbf{Parameters}           & \textbf{Top-1} & \textbf{Top-2} & \textbf{Top-3} & \textbf{Top-5}  \\ \hline
\multicolumn{1}{c|}{\multirow{4}{*}{\rotatebox[origin=c]{90}{\textbf{GCL}}}} & GBT(\cite{bielak2021graph})    & 0.71M     & 0.524	& 0.612	& 0.703	& 0.794                        \\
\multicolumn{1}{c|}{}                                          & GRACE(\cite{zhu2020deep})         & 0.74M    & 0.602	 & 0.639	& 0.726	& 0.784              \\
\multicolumn{1}{c|}{}                                          & BGRL(\cite{thakoor2021bootstrapped})          & 0.69M    & 0.589	 & 0.647	& 0.705	 & 0.739                       \\
\multicolumn{1}{c|}{}                                          & InfoG(\cite{sun2019infograph})        & 0.68M    & 0.572	 & 0.643	& 0.717	& 0.767                 \\
\hline
\multicolumn{1}{c|}{\multirow{15}{*}{\rotatebox[origin=c]{90}{\textbf{Graph Convolution Networks}}}}        & APPNP(\cite{klicpera2018predict})           & 0.74M    & 0.625	& 0.726	& 0.825	& 0.855  \\
\multicolumn{1}{c|}{}                                          & AGNN(\cite{thekumparampil2018attention})  & 0.52M    & 0.533	 & 0.745	& 0.851	& 0.954                        \\
\multicolumn{1}{c|}{}                                          & ARMA(\cite{bianchi2021graph})       & 0.45M    & 0.549	& 0.763	& 0.857	& 0.937                         \\
\multicolumn{1}{c|}{}                                          & DNA(\cite{fey2019just})        & 0.84M    & 0.607	& 0.683	& 0.772	& 0.913                      \\
\multicolumn{1}{c|}{}                                          & GAT(\cite{velivckovic2017graph})      & 0.63M    & 0.524	& 0.689	& 0.814	& 0.926                         \\
\multicolumn{1}{c|}{}                                          & GGC(\cite{li2015gated})          & 0.81M    & 0.617	& 0.813	& 0.845	& 0.951  \\
\multicolumn{1}{c|}{}                                          & GC(\cite{morris2019weisfeiler})         & 0.59M     & 0.606	& 0.769	& 0.913	& 0.962                         \\
\multicolumn{1}{c|}{}                                          & GCN2C(\cite{chen})   & 0.62M    & 0.703	& 0.829	& 0.874	& 0.957
             &            \\
\multicolumn{1}{c|}{}                                          & CC(\cite{defferrard2016convolutional})     & 0.50M    & 0.566	& 0.782	& 0.847	& 0.913                                                \\ 
\multicolumn{1}{c|}{}                                          & GUNet(\cite{gao2019graph})        & 0.96M    & 0.635	 & 0.746	& 0.873	& 0.928                         \\
\multicolumn{1}{c|}{}                                          & MPNN(\cite{gilmer2017neural})       & 0.52M    & 0.662	& 0.825	& 0.896	& 0.972                         \\
\multicolumn{1}{c|}{}                                          & RGGC(\cite{bresson2017residual})          & 0.66M    & 0.658	 & 0.744	& 0.906	 & 0.947                        \\
\multicolumn{1}{c|}{}                                          & SGAT(\cite{kim2022find})  & 0.55M    & 0.592	 & 0.694	& 0.892	& 0.955  \\
\multicolumn{1}{c|}{}                                          & TAGC(\cite{du2017topology})         & 0.57M    & 0.637	& 0.753	 & 0.827	& 0.962                         \\
\hline
\multicolumn{1}{c|}{}                                          & \textbf{Vision-HgNN}      & 0.91M    &    \textbf{0.819}	& \textbf{0.864}	& \textbf{0.942}	& \textbf{0.994}     &                     \\ \hline
\end{tabular}}}
\vspace{1mm}

\end{table}

\vspace{-9mm}
\begin{table}[htbp]
\centering
\renewcommand{\arraystretch}{1.0}
\setlength{\tabcolsep}{2.5pt}
\resizebox{0.75\textwidth}{!}{%
\subfloat{%
\begin{tabular}{cc|cccccc}
\hline
\multicolumn{2}{c|}{\textbf{Algorithms}}                      &\textbf{Parameters}                & \textbf{Top-1} & \textbf{Top-2} & \textbf{Top-3} & \textbf{Top-5}  \\ \hline
\multicolumn{1}{c|}{\multirow{6}{*}{\rotatebox[origin=c]{90}{\textbf{VCL}}}} & Barlowtwins(\cite{zbontar2021barlow})  & 8.99M    & 0.176	 & 0.264	 & 0.337	 & 0.449                         \\
\multicolumn{1}{c|}{}                                          & SimCLR(\cite{chen2020simple})       & 8.73M    & 0.189	& 0.243	& 0.408	& 0.475                       \\
\multicolumn{1}{c|}{}                                          & Byol(\cite{grill2020bootstrap})         & 8.86M   & 0.163	& 0.245	& 0.323	& 0.437                         \\
\multicolumn{1}{c|}{}                                          & Moco(\cite{he2020momentum})         & 8.73M   & 0.174	& 0.196	& 0.264	& 0.468                         \\
\multicolumn{1}{c|}{}                                          & Nnclr(\cite{dwibedi2021little})       & 9.12M   & 0.153	& 0.271	& 0.441	& 0.538                         \\
\multicolumn{1}{c|}{}                                          & SimSiam(\cite{chen2021exploring})      & 9.01M   & 0.196	& 0.301	& 0.416	& 0.561                          \\ \hline
\multicolumn{1}{c|}{}                                          & \textbf{Vision-HgNN}     &  0.91M    &    \textbf{0.819}	& \textbf{0.864}	& \textbf{0.942}	& \textbf{0.994}              &                     \\ \hline
\end{tabular}}
\qquad
}
\setlength{\tabcolsep}{2pt}
\label{tab:table1}
\vspace{-2mm}
\caption{\label{tab:cstudy}Comparative study of our proposed method and the baseline algorithms.}
\end{table}

\vspace{-9mm}
\begin{table}[htbp]
\centering
\renewcommand{\arraystretch}{1.0}
\setlength{\tabcolsep}{2.5pt}
\resizebox{0.75\textwidth}{!}{%
\qquad
\subfloat{%
\begin{tabular}{cc|cccccc}
\hline
\multicolumn{2}{c|}{\textbf{Algorithms}}                           &\textbf{Parameters}           & \textbf{Top-1} & \textbf{Top-2} & \textbf{Top-3} & \textbf{Top-5}  \\ \hline
\multicolumn{1}{c|}{\multirow{6}{*}{\rotatebox[origin=c]{90}{\textbf{ViTs}}}}        & CVT(\cite{CVT})   & 0.26M    & 0.551	& 0.764	& 0.843	& 0.965               \\
\multicolumn{1}{c|}{}                                          & CrossViT(\cite{Crossvit})     & 0.84M    & 0.493	& 0.738	& 0.842	& 0.963
                      \\
\multicolumn{1}{c|}{}                                          & ATS(\cite{fayyaz2021ats})          & 3.26M     & 0.536	& 0.725	& 0.805	& 0.942                       \\
\multicolumn{1}{c|}{}                                          & DeepViT(\cite{Deepvit})      & 3.26M    & 0.537	& 0.762	& 0.893	& 0.961
                         \\
\multicolumn{1}{c|}{}                                          & Distallation(\cite{Distillation}) & 2.06M  & 0.524	& 0.733	& 0.859	& 0.953                       \\
\multicolumn{1}{c|}{}                                          & PiT(\cite{PiT})          & 4.48M     & 0.547	& 0.716	& 0.845	& 0.962
             &            \\
\hline
\multicolumn{1}{c|}{}                                          & \textbf{Vision-HgNN}     &  0.91M    &    \textbf{0.819}	& \textbf{0.864}	& \textbf{0.942}	& \textbf{0.994}  &                     \\ \hline
\end{tabular}}}
\setlength{\tabcolsep}{2pt}
\label{tab:table1}
\vspace{-2mm}
\caption{\label{tab:cstudy}Comparative study of our proposed method and the baseline algorithms.}
\end{table}

\newpage
\clearpage

performance with a high Top-1 accuracy score of 81.9$\%$ and a Top-5 score of 99.4$\%$. The Vision-HgNN model brings a significant relative improvement of 21.87$\%$ and 16.50$\%$ in the Top-1 scores compared to the next-best baseline models T2TViT(\cite{T2TViT}) among ViTs and GCN2C(\cite{chen}) among GNNs. The ablation studies, hyperparameters optimization, and other additional experimental results are reported and discussed in the appendix.

\vspace{-3mm}
\section{Conclusion} %
\vspace{-4mm}
The challenge associated with the design of chips smaller than 7 nanometers is the increased complexity of the manufacturing process. As feature sizes shrink, the tolerance for errors in the manufacturing process decreases, making it more difficult to produce high-quality chips with consistent performance. In addition, the smaller feature sizes can lead to increased variability in the performance of the transistors, which can negatively impact the overall performance and reliability of the chip. Indeed, state-of-the-art imaging and analysis techniques are crucial in the development of next-generation semiconductor devices with feature sizes of 7nm or smaller. These techniques play a prominent role in the fabrication, inspection, and testing processes and are essential for driving the development of advanced microelectronics technologies. The high-resolution imaging of the device structures and materials allows for the identification of potential defects and deviations from design specifications, which can then be addressed through process optimization or design modification. In addition, these challenges present significant opportunities for innovation and the development of automatic material characterization methods for electron micrographs, which is essential for ensuring the quality and reliability of semiconductor devices. We conduct the first comprehensive study of the hypergraph-neural networks for electron micrograph classification tasks to improve the accuracy and efficiency of material characterization in various applications. We learn the optimal visual hypergraph structure to capture complex relationships and interactions between different spatial regions in an electron micrograph, allowing for a more robust and accurate representation of the micrograph. Hypergraph Neural Networks(HgNNs) operate on visual hypergraphs, where the patches in the micrograph are represented as hypernodes. The hyperedges in the hypergraph represent relationships between multiple patches, allowing for the capture of higher-order relationships between the patches in the micrograph. The experimental results corroborate our approach of augmenting HgAT layer stacks with a subsequent fully-connected transformer module(HgT) to achieve better performance compared to the state-of-art methods on automatic electron micrographs classification tasks. For future work,  we would endeavor to generalize our framework on other electron micrograph datasets like REM, TEM, FE-SEM, STEM, etc., for anomaly detection, segmentation, etc.

\vspace{-3mm}
\bibliography{iclr2023_conference}

\begin{thebibliography}{87}
\providecommand{\natexlab}[1]{#1}
\providecommand{\url}[1]{\texttt{#1}}
\expandafter\ifx\csname urlstyle\endcsname\relax
  \providecommand{\doi}[1]{doi: #1}\else
  \providecommand{\doi}{doi: \begingroup \urlstyle{rm}\Url}\fi

\bibitem[al.(2022{\natexlab{a}})]{neelayvformer}
Neelay~Shahet al.
\newblock Vformer: A modular pytorch library for vision transformers.
\newblock \emph{GitHub. Note: https://github.com/SforAiDl/vformer},
  2022{\natexlab{a}}.

\bibitem[al.(2022{\natexlab{b}})]{philvformer}
Phil~Wang al.
\newblock Vision transformer - pytorch.
\newblock \emph{GitHub. Note: https://github.com/lucidrains/vit-pytorch},
  2022{\natexlab{b}}.

\bibitem[Alon \& Yahav(2020)Alon and Yahav]{alon2020bottleneck}
Uri Alon and Eran Yahav.
\newblock On the bottleneck of graph neural networks and its practical
  implications.
\newblock \emph{arXiv preprint arXiv:2006.05205}, 2020.

\bibitem[Aversa et~al.(2018)Aversa, Modarres, Cozzini, Ciancio, and
  Chiusole]{aversa2018first}
Rossella Aversa, Mohammad~Hadi Modarres, Stefano Cozzini, Regina Ciancio, and
  Alberto Chiusole.
\newblock The first annotated set of scanning electron microscopy images for
  nanoscience.
\newblock \emph{Scientific data}, 5\penalty0 (1):\penalty0 1--10, 2018.

\bibitem[Ba et~al.(2016)Ba, Kiros, and Hinton]{ba2016layer}
Jimmy~Lei Ba, Jamie~Ryan Kiros, and Geoffrey~E Hinton.
\newblock Layer normalization.
\newblock \emph{arXiv preprint arXiv:1607.06450}, 2016.

\bibitem[Baek et~al.(2021)Baek, Kang, and Hwang]{baek2021accurate}
Jinheon Baek, Minki Kang, and Sung~Ju Hwang.
\newblock Accurate learning of graph representations with graph multiset
  pooling.
\newblock \emph{arXiv preprint arXiv:2102.11533}, 2021.

\bibitem[Bianchi et~al.(2021)Bianchi, Grattarola, Livi, and
  Alippi]{bianchi2021graph}
Filippo~Maria Bianchi, Daniele Grattarola, Lorenzo Livi, and Cesare Alippi.
\newblock Graph neural networks with convolutional arma filters.
\newblock \emph{IEEE transactions on pattern analysis and machine
  intelligence}, 2021.

\bibitem[Bielak et~al.(2021)Bielak, Kajdanowicz, and Chawla]{bielak2021graph}
Piotr Bielak, Tomasz Kajdanowicz, and Nitesh~V Chawla.
\newblock Graph barlow twins: A self-supervised representation learning
  framework for graphs.
\newblock \emph{arXiv preprint arXiv:2106.02466}, 2021.

\bibitem[Bresson \& Laurent(2017)Bresson and Laurent]{bresson2017residual}
Xavier Bresson and Thomas Laurent.
\newblock Residual gated graph convnets.
\newblock \emph{arXiv preprint arXiv:1711.07553}, 2017.

\bibitem[Brody et~al.(2021)Brody, Alon, and Yahav]{brody2021attentive}
Shaked Brody, Uri Alon, and Eran Yahav.
\newblock How attentive are graph attention networks?
\newblock \emph{arXiv preprint arXiv:2105.14491}, 2021.

\bibitem[Chen et~al.(2021{\natexlab{a}})Chen, Panda, and Fan]{Regionvit}
Chun-Fu Chen, Rameswar Panda, and Quanfu Fan.
\newblock Regionvit: Regional-to-local attention for vision transformers.
\newblock \emph{arXiv preprint arXiv:2106.02689}, 2021{\natexlab{a}}.

\bibitem[Chen et~al.(2021{\natexlab{b}})Chen, Fan, and Panda]{Crossvit}
Chun-Fu~Richard Chen, Quanfu Fan, and Rameswar Panda.
\newblock Crossvit: Cross-attention multi-scale vision transformer for image
  classification.
\newblock In \emph{Proceedings of the IEEE/CVF International Conference on
  Computer Vision}, pp.\  357--366, 2021{\natexlab{b}}.

\bibitem[Chen et~al.(2020{\natexlab{a}})Chen, Wei, Huang, Ding, and Li]{chen}
Ming Chen, Zhewei Wei, Zengfeng Huang, Bolin Ding, and Yaliang Li.
\newblock Simple and deep graph convolutional networks.
\newblock In \emph{International Conference on Machine Learning}, pp.\
  1725--1735. PMLR, 2020{\natexlab{a}}.

\bibitem[Chen et~al.(2020{\natexlab{b}})Chen, Kornblith, Norouzi, and
  Hinton]{chen2020simple}
Ting Chen, Simon Kornblith, Mohammad Norouzi, and Geoffrey Hinton.
\newblock A simple framework for contrastive learning of visual
  representations.
\newblock In \emph{International conference on machine learning}, pp.\
  1597--1607. PMLR, 2020{\natexlab{b}}.

\bibitem[Chen \& He(2021)Chen and He]{chen2021exploring}
Xinlei Chen and Kaiming He.
\newblock Exploring simple siamese representation learning.
\newblock In \emph{Proceedings of the IEEE/CVF Conference on Computer Vision
  and Pattern Recognition}, pp.\  15750--15758, 2021.

\bibitem[d'Ascoli et~al.(2021)d'Ascoli, Touvron, Leavitt, Morcos, Biroli, and
  Sagun]{ConViT}
St{\'e}phane d'Ascoli, Hugo Touvron, Matthew Leavitt, Ari Morcos, Giulio
  Biroli, and Levent Sagun.
\newblock Convit: Improving vision transformers with soft convolutional
  inductive biases.
\newblock \emph{arXiv preprint arXiv:2103.10697}, 2021.

\bibitem[Defferrard et~al.(2016)Defferrard, Bresson, and
  Vandergheynst]{defferrard2016convolutional}
Micha{\"e}l Defferrard, Xavier Bresson, and Pierre Vandergheynst.
\newblock Convolutional neural networks on graphs with fast localized spectral
  filtering.
\newblock \emph{Advances in neural information processing systems}, 29, 2016.

\bibitem[Deng \& Hooi(2021)Deng and Hooi]{deng2021graph}
Ailin Deng and Bryan Hooi.
\newblock Graph neural network-based anomaly detection in multivariate time
  series.
\newblock In \emph{Proceedings of the AAAI Conference on Artificial
  Intelligence}, volume~35, pp.\  4027--4035, 2021.

\bibitem[Deshpande et~al.(2020)Deshpande, Minai, and Kumar]{deshpande2020one}
Aditya~M Deshpande, Ali~A Minai, and Manish Kumar.
\newblock One-shot recognition of manufacturing defects in steel surfaces.
\newblock \emph{Procedia Manufacturing}, 48:\penalty0 1064--1071, 2020.

\bibitem[Devlin et~al.(2018)Devlin, Chang, Lee, and Toutanova]{devlin2018bert}
Jacob Devlin, Ming-Wei Chang, Kenton Lee, and Kristina Toutanova.
\newblock Bert: Pre-training of deep bidirectional transformers for language
  understanding.
\newblock \emph{arXiv preprint arXiv:1810.04805}, 2018.

\bibitem[Dosovitskiy et~al.(2020)Dosovitskiy, Beyer, Kolesnikov, Weissenborn,
  Zhai, Unterthiner, Dehghani, Minderer, Heigold, Gelly,
  et~al.]{dosovitskiy2020image}
Alexey Dosovitskiy, Lucas Beyer, Alexander Kolesnikov, Dirk Weissenborn,
  Xiaohua Zhai, Thomas Unterthiner, Mostafa Dehghani, Matthias Minderer, Georg
  Heigold, Sylvain Gelly, et~al.
\newblock An image is worth 16x16 words: Transformers for image recognition at
  scale.
\newblock \emph{arXiv preprint arXiv:2010.11929}, 2020.

\bibitem[Du et~al.(2017)Du, Zhang, Wu, Moura, and Kar]{du2017topology}
Jian Du, Shanghang Zhang, Guanhang Wu, Jos{\'e}~MF Moura, and Soummya Kar.
\newblock Topology adaptive graph convolutional networks.
\newblock \emph{arXiv preprint arXiv:1710.10370}, 2017.

\bibitem[Dwibedi et~al.(2021)Dwibedi, Aytar, Tompson, Sermanet, and
  Zisserman]{dwibedi2021little}
Debidatta Dwibedi, Yusuf Aytar, Jonathan Tompson, Pierre Sermanet, and Andrew
  Zisserman.
\newblock With a little help from my friends: Nearest-neighbor contrastive
  learning of visual representations.
\newblock In \emph{Proceedings of the IEEE/CVF International Conference on
  Computer Vision}, pp.\  9588--9597, 2021.

\bibitem[et~al.(2020)]{susmelj2020lightly}
Igor~Susmelj et~al.
\newblock Lightly.
\newblock \emph{GitHub. Note: https://github.com/lightly-ai/lightly}, 2020.

\bibitem[Fayyaz et~al.(2021)Fayyaz, Kouhpayegani, Jafari, Sommerlade, Joze,
  Pirsiavash, and Gall]{fayyaz2021ats}
Mohsen Fayyaz, Soroush~Abbasi Kouhpayegani, Farnoush~Rezaei Jafari, Eric
  Sommerlade, Hamid Reza~Vaezi Joze, Hamed Pirsiavash, and Juergen Gall.
\newblock Ats: Adaptive token sampling for efficient vision transformers.
\newblock \emph{arXiv preprint arXiv:2111.15667}, 2021.

\bibitem[Feng et~al.(2019)Feng, You, Zhang, Ji, and Gao]{feng2019hypergraph}
Yifan Feng, Haoxuan You, Zizhao Zhang, Rongrong Ji, and Yue Gao.
\newblock Hypergraph neural networks.
\newblock In \emph{Proceedings of the AAAI conference on artificial
  intelligence}, volume~33, pp.\  3558--3565, 2019.

\bibitem[Fey(2019)]{fey2019just}
Matthias Fey.
\newblock Just jump: Dynamic neighborhood aggregation in graph neural networks.
\newblock \emph{arXiv preprint arXiv:1904.04849}, 2019.

\bibitem[Fey \& Lenssen(2019)Fey and Lenssen]{Fey/Lenssen/2019}
Matthias Fey and Jan~E. Lenssen.
\newblock Fast graph representation learning with {PyTorch Geometric}.
\newblock In \emph{ICLR Workshop on Representation Learning on Graphs and
  Manifolds}, 2019.

\bibitem[Gao \& Ji(2019)Gao and Ji]{gao2019graph}
Hongyang Gao and Shuiwang Ji.
\newblock Graph u-nets.
\newblock In \emph{international conference on machine learning}, pp.\
  2083--2092. PMLR, 2019.

\bibitem[Gao et~al.(2022)Gao, Feng, Ji, and Ji]{gao2022hgnn}
Yue Gao, Yifan Feng, Shuyi Ji, and Rongrong Ji.
\newblock Hgnn: General hypergraph neural networks.
\newblock \emph{IEEE Transactions on Pattern Analysis and Machine
  Intelligence}, 2022.

\bibitem[Gilmer et~al.(2017)Gilmer, Schoenholz, Riley, Vinyals, and
  Dahl]{gilmer2017neural}
Justin Gilmer, Samuel~S Schoenholz, Patrick~F Riley, Oriol Vinyals, and
  George~E Dahl.
\newblock Neural message passing for quantum chemistry.
\newblock In \emph{International conference on machine learning}, pp.\
  1263--1272. PMLR, 2017.

\bibitem[Graham et~al.(2021)Graham, El-Nouby, Touvron, Stock, Joulin,
  J{\'e}gou, and Douze]{Levit}
Benjamin Graham, Alaaeldin El-Nouby, Hugo Touvron, Pierre Stock, Armand Joulin,
  Herv{\'e} J{\'e}gou, and Matthijs Douze.
\newblock Levit: a vision transformer in convnet's clothing for faster
  inference.
\newblock In \emph{Proceedings of the IEEE/CVF International Conference on
  Computer Vision}, pp.\  12259--12269, 2021.

\bibitem[Grill et~al.(2020)Grill, Strub, Altch{\'e}, Tallec, Richemond,
  Buchatskaya, Doersch, Avila~Pires, Guo, Gheshlaghi~Azar,
  et~al.]{grill2020bootstrap}
Jean-Bastien Grill, Florian Strub, Florent Altch{\'e}, Corentin Tallec, Pierre
  Richemond, Elena Buchatskaya, Carl Doersch, Bernardo Avila~Pires, Zhaohan
  Guo, Mohammad Gheshlaghi~Azar, et~al.
\newblock Bootstrap your own latent-a new approach to self-supervised learning.
\newblock \emph{Advances in Neural Information Processing Systems},
  33:\penalty0 21271--21284, 2020.

\bibitem[Hassani et~al.(2021)Hassani, Walton, Shah, Abuduweili, Li, and
  Shi]{hassani2021escaping}
Ali Hassani, Steven Walton, Nikhil Shah, Abulikemu Abuduweili, Jiachen Li, and
  Humphrey Shi.
\newblock Escaping the big data paradigm with compact transformers.
\newblock \emph{arXiv preprint arXiv:2104.05704}, 2021.

\bibitem[He et~al.(2016)He, Zhang, Ren, and Sun]{he2016deep}
Kaiming He, Xiangyu Zhang, Shaoqing Ren, and Jian Sun.
\newblock Deep residual learning for image recognition.
\newblock In \emph{Proceedings of the IEEE conference on computer vision and
  pattern recognition}, pp.\  770--778, 2016.

\bibitem[He et~al.(2020)He, Fan, Wu, Xie, and Girshick]{he2020momentum}
Kaiming He, Haoqi Fan, Yuxin Wu, Saining Xie, and Ross Girshick.
\newblock Momentum contrast for unsupervised visual representation learning.
\newblock In \emph{Proceedings of the IEEE/CVF conference on computer vision
  and pattern recognition}, pp.\  9729--9738, 2020.

\bibitem[Heo et~al.(2021)Heo, Yun, Han, Chun, Choe, and Oh]{PiT}
Byeongho Heo, Sangdoo Yun, Dongyoon Han, Sanghyuk Chun, Junsuk Choe, and
  Seong~Joon Oh.
\newblock Rethinking spatial dimensions of vision transformers.
\newblock In \emph{Proceedings of the IEEE/CVF International Conference on
  Computer Vision}, pp.\  11936--11945, 2021.

\bibitem[Holt \& Joy(2013)Holt and Joy]{holt2013sem}
David~Basil Holt and David~C Joy.
\newblock \emph{SEM microcharacterization of semiconductors}.
\newblock Academic Press, 2013.

\bibitem[Huang et~al.(2017)Huang, Liu, Van Der~Maaten, and
  Weinberger]{huang2017densely}
Gao Huang, Zhuang Liu, Laurens Van Der~Maaten, and Kilian~Q Weinberger.
\newblock Densely connected convolutional networks.
\newblock In \emph{Proceedings of the IEEE conference on computer vision and
  pattern recognition}, pp.\  4700--4708, 2017.

\bibitem[Hwang et~al.(2021)Hwang, Thost, Dasgupta, and Ma]{hwang2021revisiting}
EunJeong Hwang, Veronika Thost, Shib~Sankar Dasgupta, and Tengfei Ma.
\newblock Revisiting virtual nodes in graph neural networks for link
  prediction.
\newblock 2021.

\bibitem[Iandola et~al.(2016)Iandola, Han, Moskewicz, Ashraf, Dally, and
  Keutzer]{iandola2016squeezenet}
Forrest~N Iandola, Song Han, Matthew~W Moskewicz, Khalid Ashraf, William~J
  Dally, and Kurt Keutzer.
\newblock Squeezenet: Alexnet-level accuracy with 50x fewer parameters and< 0.5
  mb model size.
\newblock \emph{arXiv preprint arXiv:1602.07360}, 2016.

\bibitem[Ishiguro et~al.(2019)Ishiguro, Maeda, and Koyama]{ishiguro2019graph}
Katsuhiko Ishiguro, Shin-ichi Maeda, and Masanori Koyama.
\newblock Graph warp module: an auxiliary module for boosting the power of
  graph neural networks.
\newblock \emph{arXiv preprint arXiv:1902.01020}, 2019.

\bibitem[Jang et~al.(2016)Jang, Gu, and Poole]{jang2016categorical}
Eric Jang, Shixiang Gu, and Ben Poole.
\newblock Categorical reparameterization with gumbel-softmax.
\newblock \emph{arXiv preprint arXiv:1611.01144}, 2016.

\bibitem[Jiang et~al.(2022)Jiang, Benge, and King]{jiang2022bertvision}
Siduo Jiang, Cristopher Benge, and William~Casey King.
\newblock Bertvision--a parameter-efficient approach for question answering.
\newblock \emph{arXiv preprint arXiv:2202.12210}, 2022.

\bibitem[Kim \& Oh(2022)Kim and Oh]{kim2022find}
Dongkwan Kim and Alice Oh.
\newblock How to find your friendly neighborhood: Graph attention design with
  self-supervision.
\newblock \emph{arXiv preprint arXiv:2204.04879}, 2022.

\bibitem[Klicpera et~al.(2018)Klicpera, Bojchevski, and
  G{\"u}nnemann]{klicpera2018predict}
Johannes Klicpera, Aleksandar Bojchevski, and Stephan G{\"u}nnemann.
\newblock Predict then propagate: Graph neural networks meet personalized
  pagerank.
\newblock \emph{arXiv preprint arXiv:1810.05997}, 2018.

\bibitem[Kool et~al.(2019)Kool, Van~Hoof, and Welling]{kool2019stochastic}
Wouter Kool, Herke Van~Hoof, and Max Welling.
\newblock Stochastic beams and where to find them: The gumbel-top-k trick for
  sampling sequences without replacement.
\newblock In \emph{International Conference on Machine Learning}, pp.\
  3499--3508. PMLR, 2019.

\bibitem[Krizhevsky et~al.(2017)Krizhevsky, Sutskever, and
  Hinton]{krizhevsky2017imagenet}
Alex Krizhevsky, Ilya Sutskever, and Geoffrey~E Hinton.
\newblock Imagenet classification with deep convolutional neural networks.
\newblock \emph{Communications of the ACM}, 60\penalty0 (6):\penalty0 84--90,
  2017.

\bibitem[Lee et~al.(2019)Lee, Lee, and Kang]{lee2019self}
Junhyun Lee, Inyeop Lee, and Jaewoo Kang.
\newblock Self-attention graph pooling.
\newblock In \emph{International conference on machine learning}, pp.\
  3734--3743. PMLR, 2019.

\bibitem[Lee et~al.(2021)Lee, Lee, and Song]{ViT-SD}
Seung~Hoon Lee, Seunghyun Lee, and Byung~Cheol Song.
\newblock Vision transformer for small-size datasets.
\newblock \emph{arXiv preprint arXiv:2112.13492}, 2021.

\bibitem[Li et~al.(2018)Li, Han, and Wu]{li2018deeper}
Qimai Li, Zhichao Han, and Xiao-Ming Wu.
\newblock Deeper insights into graph convolutional networks for semi-supervised
  learning.
\newblock In \emph{Thirty-Second AAAI conference on artificial intelligence},
  2018.

\bibitem[Li et~al.(2015)Li, Tarlow, Brockschmidt, and Zemel]{li2015gated}
Yujia Li, Daniel Tarlow, Marc Brockschmidt, and Richard Zemel.
\newblock Gated graph sequence neural networks.
\newblock \emph{arXiv preprint arXiv:1511.05493}, 2015.

\bibitem[Liu et~al.(2021)Liu, Lin, Cao, Hu, Wei, Zhang, Lin, and Guo]{SwinT}
Ze~Liu, Yutong Lin, Yue Cao, Han Hu, Yixuan Wei, Zheng Zhang, Stephen Lin, and
  Baining Guo.
\newblock Swin transformer: Hierarchical vision transformer using shifted
  windows.
\newblock In \emph{Proceedings of the IEEE/CVF International Conference on
  Computer Vision}, pp.\  10012--10022, 2021.

\bibitem[Modarres et~al.(2017)Modarres, Aversa, Cozzini, Ciancio, Leto, and
  Brandino]{modarres2017neural}
Mohammad~Hadi Modarres, Rossella Aversa, Stefano Cozzini, Regina Ciancio,
  Angelo Leto, and Giuseppe~Piero Brandino.
\newblock Neural network for nanoscience scanning electron microscope image
  recognition.
\newblock \emph{Scientific reports}, 7\penalty0 (1):\penalty0 1--12, 2017.

\bibitem[Morris et~al.(2019)Morris, Ritzert, Fey, Hamilton, Lenssen, Rattan,
  and Grohe]{morris2019weisfeiler}
Christopher Morris, Martin Ritzert, Matthias Fey, William~L Hamilton, Jan~Eric
  Lenssen, Gaurav Rattan, and Martin Grohe.
\newblock Weisfeiler and leman go neural: Higher-order graph neural networks.
\newblock In \emph{Proceedings of the AAAI conference on artificial
  intelligence}, volume~33, pp.\  4602--4609, 2019.

\bibitem[Ouvrard(2020)]{ouvrard2020hypergraphs}
Xavier Ouvrard.
\newblock Hypergraphs: an introduction and review.
\newblock \emph{arXiv preprint arXiv:2002.05014}, 2020.

\bibitem[Pan et~al.(2022)Pan, Hang, Sil, and Potdar]{pan2022improved}
Lin Pan, Chung-Wei Hang, Avirup Sil, and Saloni Potdar.
\newblock Improved text classification via contrastive adversarial training.
\newblock In \emph{Proceedings of the AAAI Conference on Artificial
  Intelligence}, volume~36, pp.\  11130--11138, 2022.

\bibitem[Pham et~al.(2017)Pham, Tran, Dam, and Venkatesh]{pham2017graph}
Trang Pham, Truyen Tran, Hoa Dam, and Svetha Venkatesh.
\newblock Graph classification via deep learning with virtual nodes.
\newblock \emph{arXiv preprint arXiv:1708.04357}, 2017.

\bibitem[Ramp{\'a}{\v{s}}ek \& Wolf(2021)Ramp{\'a}{\v{s}}ek and
  Wolf]{rampavsek2021hierarchical}
Ladislav Ramp{\'a}{\v{s}}ek and Guy Wolf.
\newblock Hierarchical graph neural nets can capture long-range interactions.
\newblock In \emph{2021 IEEE 31st International Workshop on Machine Learning
  for Signal Processing (MLSP)}, pp.\  1--6. IEEE, 2021.

\bibitem[Renggli et~al.(2022)Renggli, Pinto, Houlsby, Mustafa, Puigcerver, and
  Riquelme]{PatchMerger}
Cedric Renggli, Andr{\'e}~Susano Pinto, Neil Houlsby, Basil Mustafa, Joan
  Puigcerver, and Carlos Riquelme.
\newblock Learning to merge tokens in vision transformers.
\newblock \emph{arXiv preprint arXiv:2202.12015}, 2022.

\bibitem[Rozemberczki et~al.(2021)Rozemberczki, Scherer, He, Panagopoulos,
  Riedel, Astefanoaei, Kiss, Beres, , Lopez, Collignon, and
  Sarkar]{rozemberczki2021pytorch}
Benedek Rozemberczki, Paul Scherer, Yixuan He, George Panagopoulos, Alexander
  Riedel, Maria Astefanoaei, Oliver Kiss, Ferenc Beres, , Guzman Lopez, Nicolas
  Collignon, and Rik Sarkar.
\newblock {PyTorch Geometric Temporal: Spatiotemporal Signal Processing with
  Neural Machine Learning Models}.
\newblock In \emph{Proceedings of the 30th ACM International Conference on
  Information and Knowledge Management}, pp.\  4564–4573, 2021.

\bibitem[Shang et~al.(2021)Shang, Chen, and Bi]{shang2021discrete}
Chao Shang, Jie Chen, and Jinbo Bi.
\newblock Discrete graph structure learning for forecasting multiple time
  series.
\newblock \emph{arXiv preprint arXiv:2101.06861}, 2021.

\bibitem[Simonyan \& Zisserman(2014)Simonyan and Zisserman]{simonyan2014very}
Karen Simonyan and Andrew Zisserman.
\newblock Very deep convolutional networks for large-scale image recognition.
\newblock \emph{arXiv preprint arXiv:1409.1556}, 2014.

\bibitem[Sun et~al.(2019)Sun, Hoffmann, Verma, and Tang]{sun2019infograph}
Fan-Yun Sun, Jordan Hoffmann, Vikas Verma, and Jian Tang.
\newblock Infograph: Unsupervised and semi-supervised graph-level
  representation learning via mutual information maximization.
\newblock \emph{arXiv preprint arXiv:1908.01000}, 2019.

\bibitem[Szegedy et~al.(2015)Szegedy, Liu, Jia, Sermanet, Reed, Anguelov,
  Erhan, Vanhoucke, and Rabinovich]{szegedy2015going}
Christian Szegedy, Wei Liu, Yangqing Jia, Pierre Sermanet, Scott Reed, Dragomir
  Anguelov, Dumitru Erhan, Vincent Vanhoucke, and Andrew Rabinovich.
\newblock Going deeper with convolutions.
\newblock In \emph{Proceedings of the IEEE conference on computer vision and
  pattern recognition}, pp.\  1--9, 2015.

\bibitem[Thakoor et~al.(2021)Thakoor, Tallec, Azar, Munos,
  Veli{\v{c}}kovi{\'c}, and Valko]{thakoor2021bootstrapped}
Shantanu Thakoor, Corentin Tallec, Mohammad~Gheshlaghi Azar, R{\'e}mi Munos,
  Petar Veli{\v{c}}kovi{\'c}, and Michal Valko.
\newblock Bootstrapped representation learning on graphs.
\newblock In \emph{ICLR 2021 Workshop on Geometrical and Topological
  Representation Learning}, 2021.

\bibitem[Thekumparampil et~al.(2018)Thekumparampil, Wang, Oh, and
  Li]{thekumparampil2018attention}
Kiran~K Thekumparampil, Chong Wang, Sewoong Oh, and Li-Jia Li.
\newblock Attention-based graph neural network for semi-supervised learning.
\newblock \emph{arXiv preprint arXiv:1803.03735}, 2018.

\bibitem[Tolstikhin et~al.(2021)Tolstikhin, Houlsby, Kolesnikov, Beyer, Zhai,
  Unterthiner, Yung, Steiner, Keysers, Uszkoreit, et~al.]{tolstikhin2021mlp}
Ilya~O Tolstikhin, Neil Houlsby, Alexander Kolesnikov, Lucas Beyer, Xiaohua
  Zhai, Thomas Unterthiner, Jessica Yung, Andreas Steiner, Daniel Keysers,
  Jakob Uszkoreit, et~al.
\newblock Mlp-mixer: An all-mlp architecture for vision.
\newblock \emph{Advances in Neural Information Processing Systems},
  34:\penalty0 24261--24272, 2021.

\bibitem[Touvron et~al.(2021{\natexlab{a}})Touvron, Bojanowski, Caron, Cord,
  El-Nouby, Grave, Izacard, Joulin, Synnaeve, Verbeek,
  et~al.]{touvron2021resmlp}
Hugo Touvron, Piotr Bojanowski, Mathilde Caron, Matthieu Cord, Alaaeldin
  El-Nouby, Edouard Grave, Gautier Izacard, Armand Joulin, Gabriel Synnaeve,
  Jakob Verbeek, et~al.
\newblock Resmlp: Feedforward networks for image classification with
  data-efficient training.
\newblock \emph{arXiv preprint arXiv:2105.03404}, 2021{\natexlab{a}}.

\bibitem[Touvron et~al.(2021{\natexlab{b}})Touvron, Cord, Douze, Massa,
  Sablayrolles, and J{\'e}gou]{Distillation}
Hugo Touvron, Matthieu Cord, Matthijs Douze, Francisco Massa, Alexandre
  Sablayrolles, and Herv{\'e} J{\'e}gou.
\newblock Training data-efficient image transformers \& distillation through
  attention.
\newblock In \emph{International Conference on Machine Learning}, pp.\
  10347--10357. PMLR, 2021{\natexlab{b}}.

\bibitem[Touvron et~al.(2021{\natexlab{c}})Touvron, Cord, Sablayrolles,
  Synnaeve, and J{\'e}gou]{CaiT}
Hugo Touvron, Matthieu Cord, Alexandre Sablayrolles, Gabriel Synnaeve, and
  Herv{\'e} J{\'e}gou.
\newblock Going deeper with image transformers.
\newblock In \emph{Proceedings of the IEEE/CVF International Conference on
  Computer Vision}, pp.\  32--42, 2021{\natexlab{c}}.

\bibitem[Vaswani et~al.(2017)Vaswani, Shazeer, Parmar, Uszkoreit, Jones, Gomez,
  Kaiser, and Polosukhin]{vaswani2017attention}
Ashish Vaswani, Noam Shazeer, Niki Parmar, Jakob Uszkoreit, Llion Jones,
  Aidan~N Gomez, {\L}ukasz Kaiser, and Illia Polosukhin.
\newblock Attention is all you need.
\newblock \emph{Advances in neural information processing systems}, 30, 2017.

\bibitem[Veli{\v{c}}kovi{\'c} et~al.(2017)Veli{\v{c}}kovi{\'c}, Cucurull,
  Casanova, Romero, Lio, and Bengio]{velivckovic2017graph}
Petar Veli{\v{c}}kovi{\'c}, Guillem Cucurull, Arantxa Casanova, Adriana Romero,
  Pietro Lio, and Yoshua Bengio.
\newblock Graph attention networks.
\newblock \emph{arXiv preprint arXiv:1710.10903}, 2017.

\bibitem[Vinyals et~al.(2015)Vinyals, Bengio, and Kudlur]{vinyals2015order}
Oriol Vinyals, Samy Bengio, and Manjunath Kudlur.
\newblock Order matters: Sequence to sequence for sets.
\newblock \emph{arXiv preprint arXiv:1511.06391}, 2015.

\bibitem[Wang et~al.(2022)Wang, Xie, Li, Fan, Song, Liang, Lu, Luo, and
  Shao]{PVT}
Wenhai Wang, Enze Xie, Xiang Li, Deng-Ping Fan, Kaitao Song, Ding Liang, Tong
  Lu, Ping Luo, and Ling Shao.
\newblock Pvt v2: Improved baselines with pyramid vision transformer.
\newblock \emph{Computational Visual Media}, pp.\  1--10, 2022.

\bibitem[Wu et~al.(2021)Wu, Xiao, Codella, Liu, Dai, Yuan, and Zhang]{CVT}
Haiping Wu, Bin Xiao, Noel Codella, Mengchen Liu, Xiyang Dai, Lu~Yuan, and Lei
  Zhang.
\newblock Cvt: Introducing convolutions to vision transformers.
\newblock In \emph{Proceedings of the IEEE/CVF International Conference on
  Computer Vision}, pp.\  22--31, 2021.

\bibitem[Wu et~al.(2020)Wu, Pan, Long, Jiang, Chang, and
  Zhang]{wu2020connecting}
Zonghan Wu, Shirui Pan, Guodong Long, Jing Jiang, Xiaojun Chang, and Chengqi
  Zhang.
\newblock Connecting the dots: Multivariate time series forecasting with graph
  neural networks.
\newblock In \emph{Proceedings of the 26th ACM SIGKDD international conference
  on knowledge discovery \& data mining}, pp.\  753--763, 2020.

\bibitem[Xu et~al.(2018)Xu, Li, Tian, Sonobe, Kawarabayashi, and
  Jegelka]{xu2018representation}
Keyulu Xu, Chengtao Li, Yonglong Tian, Tomohiro Sonobe, Ken-ichi Kawarabayashi,
  and Stefanie Jegelka.
\newblock Representation learning on graphs with jumping knowledge networks.
\newblock In \emph{International conference on machine learning}, pp.\
  5453--5462. PMLR, 2018.

\bibitem[Yadati et~al.(2019)Yadati, Nimishakavi, Yadav, Nitin, Louis, and
  Talukdar]{yadati2019hypergcn}
Naganand Yadati, Madhav Nimishakavi, Prateek Yadav, Vikram Nitin, Anand Louis,
  and Partha Talukdar.
\newblock Hypergcn: A new method for training graph convolutional networks on
  hypergraphs.
\newblock \emph{Advances in neural information processing systems}, 32, 2019.

\bibitem[Yu et~al.(2022)Yu, Li, Yu, Li, Huang, Wang, and Liu]{ijcai2022p328}
Hongyuan Yu, Ting Li, Weichen Yu, Jianguo Li, Yan Huang, Liang Wang, and Alex
  Liu.
\newblock Regularized graph structure learning with semantic knowledge for
  multi-variates time-series forecasting.
\newblock In Lud~De Raedt (ed.), \emph{Proceedings of the Thirty-First
  International Joint Conference on Artificial Intelligence, {IJCAI-22}}, pp.\
  2362--2368. International Joint Conferences on Artificial Intelligence
  Organization, 7 2022.
\newblock \doi{10.24963/ijcai.2022/328}.
\newblock URL \url{https://doi.org/10.24963/ijcai.2022/328}.
\newblock Main Track.

\bibitem[Yuan et~al.(2021)Yuan, Chen, Wang, Yu, Shi, Jiang, Tay, Feng, and
  Yan]{T2TViT}
Li~Yuan, Yunpeng Chen, Tao Wang, Weihao Yu, Yujun Shi, Zi-Hang Jiang,
  Francis~EH Tay, Jiashi Feng, and Shuicheng Yan.
\newblock Tokens-to-token vit: Training vision transformers from scratch on
  imagenet.
\newblock In \emph{Proceedings of the IEEE/CVF International Conference on
  Computer Vision}, pp.\  558--567, 2021.

\bibitem[Zbontar et~al.(2021)Zbontar, Jing, Misra, LeCun, and
  Deny]{zbontar2021barlow}
Jure Zbontar, Li~Jing, Ishan Misra, Yann LeCun, and St{\'e}phane Deny.
\newblock Barlow twins: Self-supervised learning via redundancy reduction.
\newblock In \emph{International Conference on Machine Learning}, pp.\
  12310--12320. PMLR, 2021.

\bibitem[Zhang et~al.()Zhang, Zhang, and Tsung]{zhanggrelen}
Weiqi Zhang, Chen Zhang, and Fugee Tsung.
\newblock Grelen: Multivariate time series anomaly detection from the
  perspective of graph relational learning.

\bibitem[Zhang et~al.(2022)Zhang, Zhang, Zhao, Chen, Arik, and Pfister]{Nest}
Zizhao Zhang, Han Zhang, Long Zhao, Ting Chen, Sercan Arik, and Tomas Pfister.
\newblock Nested hierarchical transformer: Towards accurate, data-efficient and
  interpretable visual understanding.
\newblock 2022.

\bibitem[Zhou et~al.(2021)Zhou, Kang, Jin, Yang, Lian, Jiang, Hou, and
  Feng]{Deepvit}
Daquan Zhou, Bingyi Kang, Xiaojie Jin, Linjie Yang, Xiaochen Lian, Zihang
  Jiang, Qibin Hou, and Jiashi Feng.
\newblock Deepvit: Towards deeper vision transformer.
\newblock \emph{arXiv preprint arXiv:2103.11886}, 2021.

\bibitem[Zhu et~al.(2020)Zhu, Xu, Yu, Liu, Wu, and Wang]{zhu2020deep}
Yanqiao Zhu, Yichen Xu, Feng Yu, Qiang Liu, Shu Wu, and Liang Wang.
\newblock Deep graph contrastive representation learning.
\newblock \emph{arXiv preprint arXiv:2006.04131}, 2020.

\bibitem[Zhu et~al.(2021)Zhu, Xu, Liu, and Wu]{Zhu:2021tu}
Yanqiao Zhu, Yichen Xu, Qiang Liu, and Shu Wu.
\newblock {An Empirical Study of Graph Contrastive Learning}.
\newblock \emph{arXiv.org}, September 2021.

\end{thebibliography}
\bibliographystyle{iclr2023_conference}

\newpage
\appendix
\section{Appendix}

\subsection{Modules Description}

\vspace{-1mm}
\subsubsection{Hypergraph Attention Network(HgAT)}
\vspace{-1mm}
The HgAT(\cite{velivckovic2017graph, brody2021attentive}) architecture extends the traditional convolution operation on the visual hypergraphs. It combines both local and global attention mechanisms to learn robust and expressive representations of visual hypergraphs. The hypergraph encoder(HgAT) is designed to utilize the hypergraph structure($\mathbf{H}$) and feature matrix($\mathbf{X}$) to compute the hypernode embeddings $\mathbf{h}_{v_{i}} \in \mathbb{R}^{d}, \forall i \in \mathcal{V}$. We learn the optimal embeddings $\mathbf{h}_{v_{i}}$ to preserve the high-level visual content embedded in the structural characteristics and feature attributes of the hypergraph. The layer-wise HgAT operator performs local and global-neighborhood aggregation for utilizing the powerful relational inductive bias of spatial equivariance encoded by the hypergraph’s connectivity to model the fine-grained correlations explicitly between visual patches. We perform the attention-based intra-edge neighborhood aggregation for learning the latent hyperedge embeddings as,

\vspace{-1mm}
\resizebox{0.85\linewidth}{!}{
\begin{minipage}{\linewidth}
\begin{equation}
\mathbf{h}^{(\ell)}_{e_{p}} =  \sum_{z=1}^{\mathcal{Z}} \sigma \big( \hspace{-0.25mm}  \sum_{i \hspace{0.5mm}\in \hspace{0.5mm}{\mathcal{N}_{p, i}}} \hspace{-1mm}  \alpha^{(\ell, z)}_{p, i} \mathbf{W}^{(z)}_{0}\mathbf{h}^{(\ell-1, z)}_{v_{i}} \big),  \ell=1 \ldots L_{\text{HgAT}}\label{eq:hgcnn1}
\end{equation}
\end{minipage}
}

\vspace{-1mm}
where the superscript $\ell$ denotes the layer and for scenario $\ell$ = 1, $\mathbf{h}^{(0, z)}_{v_{i}} = \mathbf{x}_{v_{i}}$. $\mathbf{h}_{e_{p}} \in \mathbb{R}^{d}$ denotes the hyperedge embeddings. We compute multiple-independent embeddings in each layer with different trainable parameters and output summed-up embeddings. $\sigma$ is the sigmoid function. The attention coefficient $\alpha_{p, i}$ determines the relative importance of the hypernode $i$ incident with the hyperedge, $p$ and is computed by,

\vspace{-6mm}
\resizebox{0.885\linewidth}{!}{
\hspace{1cm}\begin{minipage}{\linewidth}
\begin{align}
\alpha^{(\ell, z)}_{p, i} &= \frac{\exp \big(e^{(\ell, z)}_{p, i}\big)}{{\textstyle \sum_{i \hspace{0.5mm}\in \hspace{0.5mm}{\mathcal{N}_{p, i}}} \exp \big(e^{(\ell, z)}_{p, i}\big)}} ; e^{(\ell, z)}_{p, i} =\operatorname{ReLU}\big(\mathbf{W}^{(z)}_{0} \mathbf{h}^{(\ell-1, z)}_{v_{i}}\big)\label{eq:hgcnn2}
\end{align}
\end{minipage}
}

\vspace{-1mm}
where $e_{p, i}$ denote the attention score. We then model the complex relations between hyperedges and hypernodes by performing the attention-based inter-edge neighborhood aggregation for learning the expressive hypernode embeddings as described by,

\vspace{-1mm}
\resizebox{0.85\linewidth}{!}{
\begin{minipage}{\linewidth}
\begin{equation}
\mathbf{h}^{(\ell)}_{v_{i}}=\sum_{z=1}^{\mathcal{Z}} \operatorname{ReLU}\big(\mathbf{W}^{(z)}_{0}\mathbf{h}^{(\ell-1, z)}_{v_{i}} + \sum_{p \in \mathcal{N}_{i, p}} \beta^{(\ell, z)}_{i, p} \mathbf{W}^{(z)}_{1} \mathbf{h}^{(\ell, z)}_{e_{p}}\big),  \ell=1 \ldots L_{\text{HgAT}}
\label{eq:hgcnn3}
\end{equation}
\end{minipage}
} 

\vspace{-1mm}
where \resizebox{.17\textwidth}{!}{$\mathbf{W}^{(z)}_{0}, \mathbf{W}^{(z)}_{1} \in \mathbb{R}^{d \times d}$} are learnable weight matrices. We utilize the $\operatorname{ReLU}$ function to introduce non-linearity for updating the hypernode-level embeddings. The normalized attention scores $\beta_{i, p}$ specifies the importance of hyperedge $p$ incident with  hypernode $i$ and are computed by,

\vspace{-2mm}
\resizebox{0.885\linewidth}{!}{
\hspace{1cm}\begin{minipage}{\linewidth}
\begin{align}
\beta^{(\ell, z)}_{i, p} &=  \frac{\exp (\phi^{(\ell, z)}_{i, p})}{{\textstyle \sum_{p \hspace{0.5mm}\in \hspace{0.5mm}{\mathcal{N}_{i, p}}} \exp (\phi^{(\ell, z)}_{i, p})}} ; 
\phi^{(\ell, z)}_{i, p} = \operatorname{ReLU}\big(\mathbf{W}^{(z)}_{3} \cdot \big(\mathbf{W}^{(z)}_{2}\mathbf{h}^{(\ell-1, z)}_{v_{i}} \oplus \mathbf{W}^{(z)}_{2} \mathbf{h}^{(\ell, z)}_{e_{p}}\big)\big) \label{eq:hgcnn4}
\end{align}
\end{minipage}
} 

\vspace{-1mm}
where \resizebox{.115\textwidth}{!}{$\mathbf{W}^{(z)}_{2} \in \mathbb{R}^{d \times d}$} and \resizebox{.105\textwidth}{!}{$\mathbf{W}^{(z)}_{3} \in \mathbb{R}^{2d}$} are weight matrix and vector. $\oplus$ denotes the concatenation operator. $\phi_{i, p}$ is the unnormalized attention score. Stacking multiple layers broadens the receptive field, but performance degrades due to over-squashing(\cite{alon2020bottleneck}) and over-smoothing issues(\cite{li2018deeper}). In between each layer, we apply batch norm and dropout for regularization. We concatenate the embeddings of each HgAT layer, transform it through a linear projection, and serve as input to the HgT network.

\vspace{-2mm}
\subsubsection{Hypergraph Transformer(HgT)}
\vspace{-1mm}
The HgT operator generalizes the transformer neural networks(\cite{vaswani2017attention}) to arbitrary sparse hypergraph structures with full attention as a desired inductive bias for generalization. The permutation-invariant HgT module models the pairwise relations between all hypernodes and updates the hypernode-level embeddings by exploiting global contextual information in the visual hypergraphs. The HgT module with no structural priors acts as a drop-in replacement for various other methods of stacking multiple HgNN layers with residual connections(\cite{fey2019just}, \cite{xu2018representation}), virtual hypernode mechanisms(\cite{gilmer2017neural, pham2017graph}), or hierarchical pooling schemes(\cite{rampavsek2021hierarchical}, \cite{lee2019self}) to model the long-range correlations in the visual hypergraph. The HgT operator incentivizes learning the fine-grained relations to facilitate learning of expressive embeddings by spanning large receptive fields to effectively capture the high-level semantic information embedded in the hypergraph structure. The transformer encoder(\cite{vaswani2017attention}) consists of alternating layers of multiheaded self-attention(MSA) and MLP blocks.  We apply Layernorm(LN(\cite{ba2016layer})) for regularization and residual connections after every block. We skip the details for conciseness and to avoid notion clutter. Inspired by ResNets(\cite{he2016deep}), we add skip-connections through an initial connection strategy to relieve the vanishing gradients and over-smoothing issues.

\vspace{-2mm}
\resizebox{0.945\linewidth}{!}{
\hspace{2cm}\begin{minipage}{\linewidth}
\begin{align}
\mathbf{h^{\prime}}^{(\ell)}_{v_{i}} &=\operatorname{MSA}\big(\operatorname{LN}\big(\mathbf{h}^{(\ell-1)}_{v_{i}} + \mathbf{x}_{p}^{i} \mathbf{E} \big)\big)+\mathbf{h}^{(\ell-1)}_{v_{i}}, & &  \ell=1 \ldots L_{\text{HgT}} \\
\mathbf{z}^{\ell}_{v_{i}} &=\operatorname{MLP}\big(\operatorname{LN}\big(\mathbf{h^{\prime}}^{(\ell)}_{v_{i}}\big)\big)+\mathbf{h^{\prime}}^{(\ell)}_{v_{i}}, & &  \ell=1 \ldots L_{\text{HgT}} 
\end{align}
\end{minipage}
} 

\vspace{0mm}
We do not add position embeddings in skip-connections as HgAT operator had encoded the structural information into the hypernode embeddings. HgT overcomes the inherent information bottleneck of HgAT representational capacity for effective hypergraph summarization. It does so by learning hypernode-to-hypernode relations beyond the original sparse structure and distills the long-range information in the downstream layers to learn task-specific expressive hypergraph embeddings.

\vspace{-1mm}
\subsection{Algorithmic Architecture}
 Algorithm 1 summarizes the hypergraph structure learning(HgSL) module. While Algorithm 2 gives an overview of the Vision-HgNN framework. Our implementation utilizes the HgAT module to encode discrete visual hypergraphs to compute the hypernode embeddings, and the HgT module refines the embeddings through the self-attention mechanism. The HgRo module computes the hypergraph-level embedding to facilitate the classification task

\vspace{-2mm}
\begin{algorithm}
\SetAlgoLined
\DontPrintSemicolon
\textbf{Input:} Electron micrograph, $\mathbf{X^\prime} \in \mathbb{R}^{h \times w \times c}$, where $h$ is the height, $w$ is the width, and $c$ is the number of channels, patch size $p$, patch feature representation size ($d$)\;
\textbf{Output:} Visual hypergraph $\mathcal{G}$, incidence matrix $\mathbf{H} \in \mathbb{R}^{n \times n}$,  feature matrix $\mathbf{X} \in \mathbb{R}^{n \times d}$\;
\textbf{Model Parameters:} Patch and Position embedding matrices $\mathbf{E} \in \mathbb{R}^{\hspace{0.5mm} p^{2}c \times d}$, $\mathbf{E}_{pos} \in \mathbb{R}^{n \times d}$\;
$\mathbf{1}\textbf{:}$ reshape electron micrograph, $\mathbf{X^\prime} \in \mathbb{R}^{h \times w \times c} \rightarrow \mathbf{X^\prime} \in \mathbb{R}^{n \times p^{2}c}, \mathbf{x}_{p}^{i} \in \mathbb{R}^{\hspace{0.5mm}p^{2}c}, i = 1, \dots, n$ \; 
$\mathbf{2}\textbf{:}$ $\left[\mathbf{\mathbf{x}_{v_{i}}}, \ldots, \mathbf{x}_{v_{n}}\right] =\left[\mathbf{x}_{p}^{1}; \mathbf{x}_{p}^{2}; \cdots ; \mathbf{x}_{p}^{n}\right]\mathbf{E}, \mathbf{\mathbf{x}_{v_{i}}} \in \mathbb{R}^{d} , \hspace{1mm} i = 1, \dots, n$\tcp*{linear transformation} 
$\mathbf{3}\textbf{:}$ $\left[\mathbf{\mathbf{x}_{v_{i}}}, \ldots, \mathbf{x}_{v_{n}}\right] =\left[\mathbf{\mathbf{x}_{v_{i}}}, \ldots, \mathbf{x}_{v_{n}}\right] + \mathbf{E}_{pos}$, \hspace{-5.5mm}\tcp*{add position embeddings} 
$\mathbf{4}\textbf{:}$ $\mathbf{X}  = \left[\mathbf{\mathbf{x}_{v_{i}}}, \ldots, \mathbf{x}_{v_{n}}\right], \mathbf{X} \in \mathbb{R}^{n \times d}$ \tcp*{feature matrix} 
\For{patch $i=1,\dots,n$}{
    \For{patch $j=1,\dots,n$}{
    $d^{(p)}_{i, j} = \big(\sum_{z=1}^{d}{|\mathbf{\mathbf{x}^{(z)}_{v_{i}}} - \mathbf{\mathbf{x}^{(z)}_{v_{j}}}|^{2}}\big)^{1/2};  i \neq j$  \tcp*{$d^{(p)}_{i, j}$ denotes the distance similarity measure between a pair of hypernodes, $i$ and $j$ connected by any hyperedge p} 
    }
}
$\mathbf{5}\textbf{:}$ $\mathbf{H}_{j,p} = \mathbbm{1}\big\{j \in \operatorname{Top-K}\big(\operatorname{min}\big\{d^{(p)}_{i,j}\big\}, j \in \mathcal{V}\big)\big\} | \mathbf{H}_{i,p}=1$ \tcp*{$\operatorname{Top-K} \hspace{1mm} \text{returns the indices of the K-nearest hypernodes of i}$} %
\caption{Hypergraph structure learning} 
\label{alg:algo1}
\end{algorithm}
\newpage
\vspace{-2mm}
\begin{algorithm}
\SetAlgoLined
\DontPrintSemicolon
\textbf{Input:} Visual hypergraph $\mathcal{G}$, incidence matrix $\mathbf{H} \in \mathbb{R}^{n \times n}$,
 the number of hypergraph attention layers $L_{\text{HgAT}}$, number of transformer layers $L_{\text{HgT}}$, feature matrix $\mathbf{X} \in \mathbb{R}^{n \times d}$\;
\textbf{Output:}  model predictions of electron micrographs category, $y^{p}$\;
\textbf{Model Parameters:} $\mathbf{W}^{(z)}_{0}, \mathbf{W}^{(z)}_{1}, \mathbf{W}^{(z)}_{2}, \mathbf{W^{out}} \in \mathbb{R}^{\hspace{0.5mm} n \times d}, \mathbf{E} \in \mathbb{R}^{\hspace{0.5mm} p^{2}c \times d}$ \;
\For{layer $\ell=1,\dots,L_{\text{HgAT}}$}{ 
     evaluate $\alpha^{(\ell, z)}_{p, i}$ \tcp*{attention coefficient}%
     $\mathbf{h}^{(\ell)}_{e_{p}} =  \sum_{z=1}^{\mathcal{Z}} \sigma \bigg( \hspace{-0.25mm}  \sum_{i \hspace{0.5mm}\in \hspace{0.5mm}{\mathcal{N}_{p, i}}} \hspace{-1mm}  \alpha^{(\ell, z)}_{p, i} \mathbf{W}^{(z)}_{0}\mathbf{h}^{(\ell-1, z)}_{v_{i}} \bigg)$ \tcp*{hyperedge embeddings}%
     evaluate $\beta^{(\ell, z)}_{i, p}$ \tcp*{attention coefficient}%
     $\mathbf{h}^{(\ell)}_{v_{i}}=\sum_{z=1}^{\mathcal{Z}} \operatorname{ReLU}\bigg(\mathbf{W}^{(z)}_{0}\mathbf{h}^{(\ell-1, z)}_{v_{i}} + \sum_{p \in \mathcal{N}_{i, p}} \beta^{(\ell, z)}_{i, p} \mathbf{W}^{(z)}_{1} \mathbf{h}^{(\ell, z)}_{e_{p}}\bigg)$ \tcp*{hypernode embeddings} %
}
\vspace{-2mm}
compute $\mathbf{h}_{v_{i}} =  [\mathbf{h}^{(\ell)}_{v_{i}} || \ldots || \mathbf{h}^{(L_{\text{HgAT}})}_{v_{i}}]\mathbf{W}, \mathbf{W} \in \mathbb{R}^{d \cdot L_{\text{HgAT}} \times d}$ \;  
\For{layer $\ell=1,\dots,L_{\text{HgT}}$}{ 
     $\mathbf{h^{\prime}}^{(\ell)}_{v_{i}} =\operatorname{MSA}\left(\operatorname{LN}\left(\mathbf{h}^{(\ell-1)}_{v_{i}} + \mathbf{x}_{p}^{i} \mathbf{E} \right)\right) + \mathbf{h}^{(\ell-1)}_{v_{i}} $\tcp*{apply multi-head self-attention}
     $\mathbf{z}^{\ell}_{v_{i}} =\operatorname{MLP}\left(\operatorname{LN}\left(\mathbf{h^{\prime}}^{(\ell)}_{v_{i}}\right)\right)+\mathbf{h^{\prime}}^{(\ell)}_{v_{i}}$\tcp*{refine hypernode-level embeddings}
}
$\mathbf{z}^{L_{\text{HgT}}} = \textsc{READOUT}(\{ \mathbf{z}^{L_{\text{HgT}}}_{v_{1}}, \dots, \mathbf{z}^{L_{\text{HgT}}}_{v_{n}} \})$\tcp*{Aggregate hypernode embeddings}
$y^{p}_{i} =\text{softmax}\big(\mathbf{W^{out}}\mathbf{z}^{L_{\text{HgT}}}\big)$\tcp*{Category predictions}
\caption{Vision-HgNN framework} 
\label{alg:algo2}
\end{algorithm}
\vspace{-6mm}
\subsection{Experimental setup}\label{sec:experimental}
\vspace{-2mm}
The data pre-processing involves per electron micrograph standard intensity normalization with a mean and covariance of 0.5 across all the channels to obtain normalized electron micrographs to the [-1, 1] range. The size of each electron micrograph in the SEM dataset (\cite{aversa2018first}) is $1024\times 768\times 3$ pixels. We resize electron micrographs to obtain a relatively lower spatial resolution, $256\times 256\times 3$ pixels, and split the downscaled electron micrographs into non-overlapping uniform patches of size $32\times 32\times 3$ pixels. The total number of patches($\textit{n}$) for each electron micrograph is 64. The position embedding($\mathbf{E}_{pos}$) and patch embedding($\mathbf{E}$) have a dimensionality size($\textit{d}$) of 128.  We obtain the visual hypergraph representations of the electron micrographs through the Top-K nearest neighbor search algorithm with the optimal K value is 20. We utilize the k-fold cross-validation technique to evaluate the performance of our proposed method with $\text{k} = 10$. The training dataset comprises consecutive \enquote{k-2} of the folds. The validation and test dataset contains each with \enquote{1} in the remaining \enquote{2} folds. We implement an early stopping technique on the validation set to prevent the model from over-fitting and for model selection. The model is trained for 100 epochs to learn from the training dataset. We set the initial learning rate as $1e^{-3}$. The recommended batch size is 48. The optimal number of layers of $\text{HgAT}$ and $\text{HgT}$ operators, i.e., $L_{\text{HgAT}}$ and $L_{\text{HgT}}$, are 2. The number of multiple-independent replicas($\mathcal{Z}$) is 4 for the $\text{HgAT}$ operator. We utilize a learning rate scheduler to drop the learning rate by half if the Top-1 accuracy shows no improvement on the validation dataset for a waiting number of 10 epochs. We run the adam optimization algorithm to minimize the cross-entropy loss between the ground-truth labels and the model predictions. We train our model and baseline methods on multiple NVIDIA Tesla T4, Nvidia Tesla v100, and GeForce RTX 2080 GPUs built upon the PyTorch framework. We evaluate the model performance and report the evaluation metrics on the test dataset. Each computational experiment runs for a unique random seed. In this work, we report the ensemble average of the results obtained from five computational experiments. The experimental results reported are the average value of the different random seeds-based experimental run outputs.

\vspace{-2mm}
\subsection{Baseline Settings}
\vspace{-2mm}
We construct the visual graphs representation of the electron micrographs through the Top-K nearest neighbor search technique, where the patches had viewed as nodes and the edges model the pairwise associations between the semantic nearest-neighbor nodes. The baseline GNNs (\cite{rozemberczki2021pytorch, Fey/Lenssen/2019}) operate on the visual graphs to perform the classification task trained through the supervised learning approach. We utilize the graph contrastive learning(GCL, \cite{Zhu:2021tu}) algorithms to learn the unsupervised node embeddings. The node-level graph encoder of the GCL algorithms had modeled by the GAT (\cite{velivckovic2017graph}) algorithm. We compute the  graph-level embedding through the sum-pooling of the node-level embeddings. The random Forest(RF) algorithm utilizes the unsupervised graph-level embeddings to predict the electron micrograph categories trained through the supervised learning technique. We report the RF model classification accuracy on the holdout data to evaluate the quality of unsupervised embeddings. The baseline ConvNets operate on the regular grid of electron micrographs to perform classification tasks trained through a supervised learning approach. The baseline ViTs (\cite{philvformer, neelayvformer}) were trained on the supervised learning classification task to learn from the sequence of patches of each electron micrograph. Furthermore, we leverage visual-contrastive learning(VCL, \cite{susmelj2020lightly}) techniques, i.e., computer-vision-based self-supervised algorithms for performing contrastive learning of visual representations to report classification accuracy. We utilized the ResNet backbone architecture for feature extraction. To reduce the baseline model complexity, we set the dimensionality size($d$) of the patches to 64. We leverage the 10-fold cross-validation technique to evaluate the performance of the baseline methods. We train the baseline models for 100 epochs. The batch size is 48.

\vspace{-3mm}
\subsection{Study of Modules}
\vspace{-2mm}
We conduct detailed ablation studies to shed light on the relative contribution of modules for the improved overall performance of our framework. We gradually exclude the modules to design several variants of our framework and then investigate the variant model's performance compared to the Vision-HgNN model on the SEM dataset to demonstrate the efficacy and support the rationale of our modules. We refer to w/o HgAT and w/o HgT as Vision-HgNN models without HgAT and HgT modules, respectively. Table \ref{tab:ab} shows the results of the ablation studies.  

\vspace{-3mm}
\begin{table}[htbp]
\centering
\setlength{\tabcolsep}{4.5pt}
\label{tab:table1}
\vspace{-2mm}
\resizebox{1.1\textwidth}{!}{%
\subfloat{%
\hspace{-10mm}\begin{tabular}{cc|c|c|c|c|c|c|cc}
\hline
\multicolumn{2}{c|}{\textbf{Algorithms}}                                     & \textbf{Top-1} & \textbf{Top-2} & \textbf{Top-3} & \textbf{Top-5}  & \textbf{Avg-Precision} & \textbf{Avg-Recall} & \textbf{Avg-F1 Score}  \\ \hline
\multicolumn{1}{c}{\multirow{6}{*}{\rotatebox[origin=c]{90}{\textbf{}}}} & Vision-HgNN      & \textbf{0.819$\pm$0.005}	& \textbf{0.864$\pm$0.009}
	& \textbf{0.942$\pm$0.005}	& \textbf{0.990$\pm$0.001}	& \textbf{0.784$\pm$0.006}	& \textbf{0.718$\pm$0.008}	& \textbf{0.731$\pm$0.004} \\ \hline
\multicolumn{1}{c}{}                                          &  w/o HgAT    & 0.682$\pm$0.008 	& 0.783$\pm$0.007	& 0.900$\pm$0.013 	& 0.968$\pm$0.007	& 0.713$\pm$0.008	& 0.662$\pm$0.005	& 0.657$\pm$0.005 \\
\multicolumn{1}{c}{}                                          &    w/o HgT    & 0.497$\pm$0.013	& 0.644$\pm$0.010	& 0.834$\pm$0.008 	& 0.946$\pm$0.009 	& 0.588$\pm$0.007	& 0.543$\pm$0.002	& 0.548$\pm$0.001 \\ \hline
\end{tabular}}}
\vspace{-2mm}
\caption{\label{tab:ab}The table reports the results of ablation studies.}
\end{table}

\vspace{-3mm}

\begin{table}[htbp]
\centering
\resizebox{0.65\textwidth}{!}{%
\begin{tabular}{@{}c|ccc|c@{}}
\toprule
\multirow{2}{*}{\textbf{Category}}   & \multicolumn{3}{c|}{\textbf{Multi-class metrics}}                                    \\ \cmidrule(lr){2-4}
                            & \multicolumn{1}{c|}{\textbf{Precision}} & \multicolumn{1}{c|}{\textbf{Recall}} & \textbf{F1 Score} &                        \\ \midrule
Biological                  & \multicolumn{1}{c|}{0.700$\pm$0.016}     & \multicolumn{1}{c|}{0.728$\pm$0.004}      & 0.713$\pm$0.007 \\
Tips                        & \multicolumn{1}{c|}{0.705$\pm$0.014}     & \multicolumn{1}{c|}{0.656$\pm$0.007}      &  0.608$\pm$0.016                            \\
Fibres                      & \multicolumn{1}{c|}{0.922$\pm$0.033}     & \multicolumn{1}{c|}{0.625$\pm$0.010}      &  0.751$\pm$0.011                         \\
Porous Sponge               & \multicolumn{1}{c|}{0.833$\pm$0.009}     & \multicolumn{1}{c|}{0.852$\pm$0.009}      &  0.842$\pm$0.009                            \\
Films Coated Surface        & \multicolumn{1}{c|}{0.849$\pm$0.002}     & \multicolumn{1}{c|}{0.837$\pm$0.003}      &  0.843$\pm$0.000                    \\
Patterned surface           & \multicolumn{1}{c|}{0.766$\pm$0.007}     & \multicolumn{1}{c|}{0.823$\pm$0.020}      &  0.793$\pm$0.013                         \\
Nanowires                   & \multicolumn{1}{c|}{0.751$\pm$0.009}     & \multicolumn{1}{c|}{0.756$\pm$0.001}      &  0.754$\pm$0.004                         \\
Particles                   & \multicolumn{1}{c|}{0.816$\pm$0.017}     & \multicolumn{1}{c|}{0.481$\pm$0.055}      &  0.572$\pm$0.033                       \\
MEMS devices                & \multicolumn{1}{c|}{0.676$\pm$0.003}     & \multicolumn{1}{c|}{0.705$\pm$0.000}      &  0.670$\pm$0.013                         \\
Powder                      & \multicolumn{1}{c|}{0.822$\pm$0.013}     & \multicolumn{1}{c|}{0.721$\pm$0.002}      &  0.767$\pm$0.004
                          \\ \bottomrule
\end{tabular}}
\vspace{-1mm}
\caption{The table reports the performance of the Vision-HgNN framework on each electron-micrograph category classification task of the SEM dataset.}
\label{tab:unavg}
\end{table}

\vspace{-3mm}
We additionally report the average precision, recall, and F1 score across the electron micrograph categories to evaluate the framework performance on the highly unbalanced SEM dataset. The w/o HgAT variant model reports a drop in performance of 9.06$\%$ in the precision score, 7.79$\%$ in the recall, and 10.12$\%$ in the F1 score w.r.t. Vision-HgNN model. Likewise, regarding w.r.t precision, recall, and F1 score, we observe a 25$\%$, 24.37$\%$, and 25.03$\%$ decline in w/o HgT variant model performance compared to the Vision-HgNN model. The variant model's performance disentangles the relative gains of each module and corroborates the hypothesis of the joint optimization of modules for better learning on the visual hypergraphs. The experimental results support our modules, HgAT and HgT effectiveness for capturing the short-range and high-level, long-range correlations, respectively, for the improved overall performance of our proposed framework. Table \ref{tab:unavg} reports the unaveraged multi-class evaluation metrics such as precision, recall, and F1 score of the Vision-HgNN framework performance on each predefined electron micrograph category. The results reported in Table \ref{tab:unavg} demonstrate that our proposed framework generalizes well despite the complexity of patterns across the broad spectrum of electron micrograph categories, with a relatively higher score for the more labeled electron micrograph categories w.r.t. to the fewer labeled categories. The less electron micrograph-specific relational inductive bias offers an advantage for the Vision-HgNN framework to perform better on the classification task than the traditional methods. 

\vspace{-3mm}
\subsection{Ablation Studies}
\vspace{-2mm}
Our proposed Vision-HgNN framework consists of HgSL, HgAT, HgT, and HgRo modules. We study the impact of each module in great detail that is responsible for the enhanced  performance of our framework by substituting the modules with well-known algorithms of similar functionality to design replaced models. We compare the performance of the replaced models with our proposed framework to support the efficacy of our modules.

\vspace{-2mm}
\subsubsubsection{\textbf{Study of HgSL Module:-}}
We study the effectiveness of the HgSL module in modeling the complex relations among the spatially and semantically dependent visual patches for a structured representation of electron micrographs. The HgSL module learns the optimal K-uniform bi-directed hypergraph structure of the electron micrographs through the top-K nearest neighbor search algorithm. The most popular graph-based approaches for constructing the optimal graph structure are classified into two categories, (a) constructing a K-regular bi-directed graph or (b) the K-irregular bi-directed graph. The K-irregular graph construction techniques overcome the limitations of K-uniform graph structure learning techniques, which are not continuously differentiable. The former algorithmic approach in the literature includes (a) the top-K nearest neighbor search strategy using cosine similarity(CS,\cite{deng2021graph}) and (b) structure learning through implicit correlations of node embeddings(MTGNN, \cite{wu2020connecting}). The later algorithmic techniques include (a) parametrization of the adjacency matrix and sample through the Gumbel reparameterization trick (\cite{jang2016categorical, kool2019stochastic}) to output the link probability between nodes(\text{GPT}, \cite{shang2021discrete}), (b) regularized graph generation(RGG, \cite{ijcai2022p328}) to learn sparse implicit graph by dropping redundant connections between nodes, and (c) graph relational learning using the self-attention mechanism(GRL, \cite{zhanggrelen}) to construct a graph from observed data. We refer to the Vision-HgNN model for which the HgSL module is modeled with the different operators as follows,

\vspace{-3mm}
\begin{itemize}
\itemsep0em
\item $\textbf{w/} \hspace{1mm} \text{CS}$: $\text{Vision-HgNN}$ model with the $\text{CS}$ operator.
\item $\textbf{w/} \hspace{1mm} \text{MTGNN}$: $\text{Vision-HgNN}$ model with the $\text{MTGNN}$ operator.
\item $\textbf{w/} \hspace{1mm} \text{GPT}$: $\text{Vision-HgNN}$ model with the $\text{GPT}$ operator.
\item $\textbf{w/} \hspace{1mm} \text{RGG}$: $\text{Vision-HgNN}$ model with the $\text{RGG}$ operator.
\item $\textbf{w/} \hspace{1mm} \text{GRL}$: $\text{Vision-HgNN}$ model with the $\text{GRL}$ operator.
\end{itemize}

\vspace{-2mm}
Table \ref{tab:sbmodels1} reports the performance of the replaced models compared to the Vision-HgNN framework. The Top-1 scores of the substituted models, $\textbf{w/} \hspace{1mm} \text{MTGNN}$, $\text{w/} \hspace{1mm} \text{GPT}$, $\text{w/} \hspace{1mm} \text{RGG}$,  $\text{w/} \hspace{1mm} \text{GRL}$ declined by 10.74$\%$, 12.94$\%$, 15.63$\%$ , 12.69$\%$ on SEM dataset compared to the Vision-HgNN model. The impact of $\text{w/} \hspace{1mm} \text{CS}$ is marginal and achieves on-par performance compared to the Vision-HgNN model. The results show the advantages of utilizing the HgSL module modeled with the top-K nearest neighbor search technique, a simple
yet effective similarity learning technique based on the Euclidean distance between patches for capturing the underlying higher-order relational information in visual hypergraphs.

 \vspace{-4mm}
\begin{table}[htbp]
\centering
\setlength{\tabcolsep}{3pt}
\label{tab:table1}
\vspace{-2mm}
\resizebox{0.8\textwidth}{!}{%
\subfloat{%
\begin{tabular}{cc|c|c|c|c|c|c|cc}
\hline
\multicolumn{2}{c|}{\textbf{Algorithms}}                                     & \textbf{Top-1} & \textbf{Top-2} & \textbf{Top-3} & \textbf{Top-5}  & \textbf{Avg-Precision} & \textbf{Avg-Recall} & \textbf{Avg-F1 Score}  \\ \hline
\multicolumn{1}{c}{\multirow{6}{*}{\rotatebox[origin=c]{90}{\textbf{}}}} & Vision-HgNN      & 0.819	& 0.864	& 0.942	& 0.994	& 0.784	& 0.718	& 0.731   \\ \hline
\multicolumn{1}{c}{}                                          &  $\text{w/} \hspace{1mm} \text{CS}$    & 0.790	& 0.833	& 0.909	& 0.959	& 0.756	& 0.693	& 0.706 \\
\multicolumn{1}{c}{}                                          &  $\text{w/} \hspace{1mm} \text{MTGNN}$    & 0.731	& 0.763	& 0.843	& 0.897	& 0.699	& 0.639	& 0.644 \\ \hline
\multicolumn{1}{c}{}                                          &    $\text{w/} \hspace{1mm} \text{GPT}$    & 0.713	& 0.744	& 0.822	& 0.875	& 0.682	& 0.616	& 0.632   \\ 
\multicolumn{1}{c}{}                                          &    $\text{w/} \hspace{1mm} \text{RGG}$    & 0.691	& 0.718	& 0.788	& 0.844	& 0.668	& 0.621	& 0.627  \\ 
\multicolumn{1}{c}{}                                          &    $\text{w/} \hspace{1mm} \text{GRL}$    & 0.715	& 0.749	& 0.826	& 0.862	& 0.694	& 0.633	& 0.645  \\ \hline
\end{tabular}}}
\vspace{-1mm}
\caption{\label{tab:sbmodels1}The table reports the comparative study of the various structure learning techniques.}
\end{table}

 \vspace{-3mm}
\subsubsubsection{\textbf{Study of HgAT Module:-}}
(a) We study the importance of the \textbf{attention mechanism} in the hypernode-level hypergraph encoder(HgAT) to compute the expressive hypernode embeddings($\mathbf{h}_{v_{i}}$) of visual hypergraphs. We disable the attention mechanism of the layerwise HgAT operator. We perform the unweighted sum-pooling operation on the neural messages in the intra- and inter-neighborhood aggregation scheme for computing hypergraph embeddings. We refer to the Vision-HgNN model in the absence of the attention-mechanism for determining the hyperedge($\mathbf{h}_{e_{p}}$) and hypernode($\mathbf{h}_{v_{i}}$) embeddings as follows,

\vspace{-3mm}
\begin{itemize}
\item $\textbf{w/o} \hspace{1mm} \mathbf{\alpha_{p, i}}, \mathbf{\beta_{i, p}}$: \hspace{-1mm}$\textbf{Vision-HgNN}$ model without the attention mechanism.
\end{itemize}

\vspace{-3mm}

Eliminating the attention mechanism degrades the replaced model performance, as evident in Table \ref{tab:sbmodels2}. In particular, it decreases the Top-1 accuracy w.r.t. the Vision HgNN model by more than 13.79$\%$ on the SEM dataset. The attention mechanism automatically learns the relative importance among the incident hypernodes and hyperedges during the local and global neighborhood aggregation. It provides beneficial inductive bias to capture the dominant-visual patterns in the electron micrographs across the categories. The hypergraph attention mechanism models the fine-grained correlations for encoding the higher-order relationships in the embeddings for better learning the visual hypergraphs.

\vspace{-5mm}

\begin{table}[htbp]
\centering
\setlength{\tabcolsep}{3pt}
\label{tab:table1}
\resizebox{0.85\textwidth}{!}{%
\subfloat{%
\begin{tabular}{cc|c|c|c|c|c|c|cc}
\hline
\multicolumn{2}{c|}{\textbf{Algorithms}}                                     & \textbf{Top-1} & \textbf{Top-2} & \textbf{Top-3} & \textbf{Top-5}  & \textbf{Avg-Precision} & \textbf{Avg-Recall} & \textbf{Avg-F1 Score}  \\ \hline
\multicolumn{1}{c}{\multirow{6}{*}{\rotatebox[origin=c]{90}{\textbf{}}}} & Vision-HgNN      & 0.819	& 0.864	& 0.942	& 0.994	& 0.784	& 0.718	& 0.731     \\ \hline
\multicolumn{1}{c}{}                                          &  $\textbf{w/o} \hspace{1mm} \mathbf{\alpha_{p, i}}, \mathbf{\beta_{i, p}}$    & 0.706	& 0.742	& 0.812	& 0.827	& 0.667	& 0.619	& 0.630       \\ \hline
\end{tabular}}}
\vspace{-1mm}
\caption{\label{tab:sbmodels2}The table shows the comparative study of the model's performance with and without the attention mechanism in HgAT.}
\end{table}

 \vspace{-1mm}
 
(b) We analyze the usefulness of \textbf{latent master-hypernode} to encode the long-range hypernode relations in the visual hypergraph embeddings for enhancing classification performance. We add a superhypernode(or virtual master hypernode, \cite{gilmer2017neural, ishiguro2019graph, pham2017graph, hwang2021revisiting}) for augmenting the visual hypergraphs. The master hypernode had connected to all hyperedges of the augmented hypergraph. It provides an additional route for neural message-passing schema on visual hypergraphs. The master hypernode embeddings contain the global latent information of the visual hypergraphs. Each hyperedge reads and writes to transform the master hypernode embedding through the intra- and inter-neighborhood aggregation-based message-passing attention networks. We design a replaced model by disabling the HgT module to operate on the augmented hypergraphs as follows,

\vspace{-1mm}
\begin{itemize}
\item $\textbf{w/} \hspace{1mm} \text{Virtual}$: $\textbf{Vision-HgNN}$ model without the HgT module to operate on augmented hypergraphs.
\end{itemize}

\vspace{-1mm}
The degradation of the variant performance is evident in Table \ref{tab:sbmodelslatenthypernode}, which shows a drop of 11.72$\%$ in the Top-1 accuracy compared to the Vision HgNN model. The latent \enquote{master} hypernode is ineffective for capturing long-range hypernode-to-hypernode relations to encode the spatial dependencies among patches in the hypergraphs embeddings for effectively learning on the visual hypergraphs-topology compared to the HgT module.

\vspace{-3mm}
\begin{table}[htbp]
\centering
\setlength{\tabcolsep}{3pt}
\label{tab:table1}
\vspace{-2mm}
\resizebox{0.85\textwidth}{!}{%
\subfloat{%
\begin{tabular}{cc|c|c|c|c|c|c|cc}
\hline
\multicolumn{2}{c|}{\textbf{Algorithms}}                                     & \textbf{Top-1} & \textbf{Top-2} & \textbf{Top-3} & \textbf{Top-5}  & \textbf{Avg-Precision} & \textbf{Avg-Recall} & \textbf{Avg-F1 Score}  \\ \hline
\multicolumn{1}{c}{\multirow{6}{*}{\rotatebox[origin=c]{90}{\textbf{}}}} & Vision-HgNN      & 0.819	& 0.864	& 0.942	& 0.994	& 0.784	& 0.718	& 0.731                   \\ \hline
\multicolumn{1}{c}{}                                          &  $\text{w/} \hspace{1mm} \text{Virtual}$    & 0.723	& 0.761	& 0.832	& 0.847	& 0.684	& 0.634	& 0.648 \\ \hline
\end{tabular}}}
\vspace{-1mm}
\caption{\label{tab:sbmodelslatenthypernode}The table shows the comparative study of the model's performance with and without the latent \enquote{master} hypernode.}
\end{table}

\vspace{-5mm}

\begin{table}[htbp]
\centering
\setlength{\tabcolsep}{3pt}
\label{tab:table1}
\vspace{-2mm}
\resizebox{0.875\textwidth}{!}{%
\subfloat{%
\begin{tabular}{cc|c|c|c|c|c|c|cc}
\hline
\multicolumn{2}{c|}{\textbf{Algorithms}}                                     & \textbf{Top-1} & \textbf{Top-2} & \textbf{Top-3} & \textbf{Top-5}  & \textbf{Avg-Precision} & \textbf{Avg-Recall} & \textbf{Avg-F1 Score}  \\ \hline
\multicolumn{1}{c}{\multirow{6}{*}{\rotatebox[origin=c]{90}{\textbf{}}}} & Vision-HgNN      & 0.819	& 0.864	& 0.942	& 0.994	& 0.784	& 0.718	& 0.731                    \\ \hline
\multicolumn{1}{c}{}                                          &  $\text{w/} \hspace{1mm} \text{TopK \hspace{0.5mm}Pooling}$ & 0.695	& 0.731	& 0.803	& 0.814	& 0.657	& 0.603	& 0.620           \\ \hline
\multicolumn{1}{c}{}                                          &  $\text{w/} \hspace{1mm} \text{SAG \hspace{0.5mm}Pooling}$  & 0.732	& 0.770	& 0.843	& 0.866	& 0.699	& 0.642	& 0.656  \\ \hline
\end{tabular}}}
\vspace{-1mm}
\caption{\label{tab:sbmodelslatenthierarchical}The table shows the comparative study of the model's performance with different hierarchical pooling schemes.}
\end{table}

\vspace{-2mm}

(c) We study the effectiveness of \textbf{hierarchical pooling schemes}(\cite{gao2019graph, lee2019self}) for better learning the long-range, higher-order dependencies in visual hypergraphs. The spatial pooling operator learns the hierarchical representations in a two-stage approach to model the large portions in the visual hypergraphs. (a) assigns a score to all hypernodes and drops the low-scoring hypernodes and the corresponding hyperedges, which contain noise or less prominent patterns. (b) samples the high-scoring hypernodes to obtain hierarchical-induced visual subhypergraphs. (c) performs the hierarchical message passing schemes on pooled hypergraphs for encoding the hypergraph’s local and global structure information to compute higher-order representations of the visual hypergraphs. We obtain a replaced model by (a) disabling the HgT module, (b) stacking three layers of pooling operators(pooling ratio($p_{r}$) as 0.75) interleaved with the HgAT module, and refer to the Vision-HgNN model as follows,

\vspace{-2mm}
\begin{itemize}
\itemsep0em
\item $\textbf{w/} \hspace{1mm} \text{TopK \hspace{0.5mm}Pooling}$: \hspace{-0.25mm}$\textbf{Vision-HgNN}$ model with the $\text{TopK \hspace{0.5mm}Pooling}$ operator (\cite{gao2019graph}).
\item $\textbf{w/} \hspace{1mm} \text{SAG \hspace{0.5mm}Pooling}$: \hspace{-0.25mm}$\textbf{Vision-HgNN}$ model with the $\text{SAG \hspace{0.5mm}Pooling}$ operator(\cite{lee2019self}).
\end{itemize}

\vspace{-3mm}

The performance degradation of the replaced models was evident in the SEM dataset, as indicated in Table \ref{tab:sbmodelslatenthierarchical}. The replaced models, $\text{w/} \hspace{1mm} \text{TopK \hspace{0.5mm}Pooling}$, $\text{w/} \hspace{1mm} \text{SAG \hspace{0.5mm}Pooling}$ yields an overall 15.14$\%$, 10.62$\%$ relative lower Top-1 accuracy compared to the Vision-HgNN model. The self-attention mechanism-based HgT module is more effective for capturing long-range spatial dependencies by computing all pairwise interactions among hypernodes through a position-agnostic fashion for learning about visual hypergraphs. 

\vspace{-2mm}

\subsubsubsection{\textbf{Study of HgT Module:-}} 
Inspired by using a special-classification token($<\hspace{-1mm}\text{CLS}\hspace{-1mm}>$) in transformer architectures (\cite{jiang2022bertvision, pan2022improved, devlin2018bert}) for sentence-level classification tasks. We will append a special-patch($<\hspace{-1mm}\text{CLS}\hspace{-1mm}>$) embedding along with the hypernode's embeddings corresponding to the visual hypergraphs as the input set of embeddings sequence to the HgT module. Note: We do not add the special patch($<\hspace{-1mm}\text{CLS}\hspace{-1mm}>$) as a virtual hypernode to the visual hypergraph. This method overcomes the drawbacks of the information bottleneck problems of the virtual hypernode augmentation technique. The former algorithm prevents learning pairwise relationships between hypernodes except with the virtual hypernode. The special-patch incorporated variant model encodes the one-to-one relations between the $<\hspace{-1mm}\text{CLS}\hspace{-1mm}>$ patch and every other patch of the visual hypergraphs in the hypergraph embeddings with the self-attention module. We hypothesize that the special-patch($<\hspace{-1mm}\text{CLS}\hspace{-1mm}>$) embedding contains complete visual hypergraph information. We apply the softmax operator to special-patch embedding and predict the category of the visual hypergraphs. Thus, we obtain a replaced model with the special-patch-inspired readout mechanism coupled with the HgT module as follows,

\vspace{-1mm}
\begin{itemize}
\itemsep0em
\item $\text{w/} \hspace{1mm} <\hspace{-1mm}\text{CLS}\hspace{-1mm}>$: The $\text{Vision-HgNN}$ model with $<\hspace{-1mm}\text{CLS}\hspace{-1mm}>$ patch in $\text{HgT}$ module and disabled HgRo module. 
\end{itemize}

\vspace{-5mm}
\begin{table}[htbp]
\centering
\setlength{\tabcolsep}{3pt}
\label{tab:table1}
\vspace{-2mm}
\resizebox{0.825\textwidth}{!}{%
\subfloat{%
\begin{tabular}{cc|c|c|c|c|c|c|cc}
\hline
\multicolumn{2}{c|}{\textbf{Algorithms}}                                     & \textbf{Top-1} & \textbf{Top-2} & \textbf{Top-3} & \textbf{Top-5}  & \textbf{Avg-Precision} & \textbf{Avg-Recall} & \textbf{Avg-F1 Score}  \\ \hline
\multicolumn{1}{c}{\multirow{6}{*}{\rotatebox[origin=c]{90}{\textbf{}}}} & Vision-HgNN      & 0.819	& 0.864	& 0.942	& 0.994	& 0.784	& 0.718	& 0.731                      \\ \hline
\multicolumn{1}{c}{}                                          &  $\text{w/} \hspace{1mm} <\hspace{-1mm}\text{CLS}\hspace{-1mm}>$    & 0.747	& 0.779	& 0.839	& 0.917	& 0.707	& 0.672	& 0.677  \\   \hline
\end{tabular}}}
 \vspace{-1mm}
\caption{\label{tab:sbmodelsHgT}The table reports the comparative study of the special-token-based hypergraph readout mechanism with our proposed method.}
\end{table} 

\vspace{-3mm}
The experimental results had illustrated in Table \ref{tab:sbmodelsHgT}. We notice a drop in the variant performance compared to our proposed method. The Top-1 accuracy of the design variant, $\text{w/} \hspace{1mm} <\hspace{-1mm}\text{CLS}\hspace{-1mm}>$, declined by 8.79$\%$ on the SEM dataset compared to the Vision-HgNN model. 

\vspace{-2mm}
\begin{table}[htbp]
\centering
\setlength{\tabcolsep}{5pt}
\label{tab:table1}
\vspace{-2mm}
\resizebox{0.9\textwidth}{!}{%
\subfloat{%
\begin{tabular}{cc|c|c|c|c|c|c|cc}
\hline
\multicolumn{2}{c|}{\textbf{Algorithms}}                                     & \textbf{Top-1} & \textbf{Top-2} & \textbf{Top-3} & \textbf{Top-5}  & \textbf{Avg-Precision} & \textbf{Avg-Recall} & \textbf{Avg-F1 Score}  \\ \hline
\multicolumn{1}{c}{\multirow{6}{*}{\rotatebox[origin=c]{90}{\textbf{}}}} & Vision-HgNN      & 0.819	& 0.864	& 0.942	& 0.994	& 0.784	& 0.718	 & 0.731                \\ \hline
\multicolumn{1}{c}{}                                          &  $\text{w/}  \text{GMT}$    & 0.774	& 0.816	& 0.890	& 0.939	& 0.741	& 0.679	& 0.691            \\
\multicolumn{1}{c}{}                                          &  $\text{w/}  \text{GA}$    & 0.786	& 0.828	& 0.884	& 0.954	& 0.752	& 0.703	& 0.716           \\ 
\multicolumn{1}{c}{}                                          &    $\text{w/} \text{Set2Set}$    &  0.719	& 0.741	& 0.818	& 0.882	& 0.698	& 0.650	& 0.658      \\ 
\multicolumn{1}{c}{}                                          &    $\text{w/} \text{GSM}$    &  0.737	& 0.753	& 0.841	& 0.909	& 0.716	& 0.666	& 0.675     \\  \hline
\end{tabular}}}
\vspace{-1mm}
\caption{\label{tab:sbmodelsHgRo}The table reports the comparative study of the hypergraph readout baseline operators.}
\end{table}

\vspace{-3mm}

\subsubsubsection{\textbf{Study of HgRo Module:-}}
We probe the HgRo module's effectiveness compared to the well-known algorithms of identical functionality. The hypergraph readout module performs global average pooling on the hypernode-level embeddings to compute the visual hypergraph-level embedding. We utilize well-known methods as baseline operators to perform global-pooling operations on the hypergraphs. The list includes GraphMultisetTransformer(GMT, \cite{baek2021accurate}), GlobalAttention(GA, \cite{li2015gated}), Set2Set(\cite{vinyals2015order}), and Global Summation Pooling(GSM). We refer to the $\textbf{Vision-HgNN}$ model with the baseline operators for modeling the global readout function(HgRo module) as follows:

\vspace{-2mm}
\begin{itemize}
\itemsep0em
\item $\text{w/} \hspace{1mm} \text{GMT}$: The $\text{Vision-HgNN}$ model with $\text{GMT}$ operator.
\item $\text{w/} \hspace{1mm} \text{GA}$: The $\text{Vision-HgNN}$ model with $\text{GA}$ operator.
\item $\text{w/} \hspace{1mm} \text{Set2Set}$: The $\text{Vision-HgNN}$ model with $\text{Set2Set}$ operator.
\item $\text{w/} \hspace{1mm} \text{GSM}$: The $\text{Vision-HgNN}$ model with $\text{GSM}$ operator.
\end{itemize}

\vspace{-2mm}

The results reported in Table ~\ref{tab:sbmodelsHgRo} across all the evaluation metrics demonstrate no significant improvements in the replaced model's performance compared to our proposed method with the global average pooling operator. Overall, our hypergraph readout module proves effective by learning to compute the optimal hypergraph-level representations while maximally preserving the visual hypergraph's global-contextual information.

\vspace{-3mm}

\subsection{Hyperparameter Studies}
\vspace{-2mm}

We perform an in-depth hyperparameter tuning to determine the optimal hyperparameters of our framework. The algorithm hyperparameters are (1) the dimensionality of embedding($d$). (2) The stack of HgAT operators($L_{\text{HgAT}}$) and the HgT operators($L_{\text{HgT}}$). Hyperparameters had chosen from the following ranges:  embedding dimension(d) $\in [32, 256]$,  $L_{\text{HgAT}}$ $\in [1, 8]$, and $L_{\text{HgT}}$ $\in [1, 6]$. We perform hyperparameter optimization using the grid-search technique to yield the optimal performance of our proposed method on the validation dataset in terms of the Top-1 classification accuracy. For each experiment, we change the hyperparameter under investigation to determine the impact on the model performance. The optimal hyperparameters determined from the study are as follows, d is 128, $L_{\text{HgAT}}$, and $L_{\text{HgAT}}$ is 2.

\vspace{-4mm}
\begin{table}[htbp]
\centering
\setlength{\tabcolsep}{2pt}
\resizebox{0.9\textwidth}{!}{%
\subfloat{%
\begin{tabular}{@{}c|c|c|c|cc@{}}
\hline
$d$ & $32$ & $64$  & $128$ & $256$ \\
\hline
 &  0.632  & 0.740  & 0.813  &  0.802  \\
\hline
\end{tabular}}
\qquad
\subfloat{%
\begin{tabular}{@{}c|c|c|c|cc@{}}
\hline
$L_{\text{HgAT}}$ &  $1$ & $2$  & $4$ & $8$ \\
\hline
  & 0.712  & 0.813  & 0.673  & 0.520  \\
\hline
\end{tabular}}
\qquad
\subfloat{%
\begin{tabular}{@{}c|c|c|c|cc@{}}
\hline
$L_{\text{HgT}}$ &  $1$ & $2$  & $4$ & $6$ \\
\hline  
  & 0.754  & 0.813   & 0.830  & 0.798   \\
\hline
\end{tabular}}}
\caption{The table reports the experimental results of the hyperparameter study.}
\label{table:Hs}
\end{table}

\vspace{-3mm}
\subsection{SEM Dataset}
\begin{figure}[htbp]

     \subfloat{\hspace{-10mm}\includegraphics[width=0.22\textwidth]{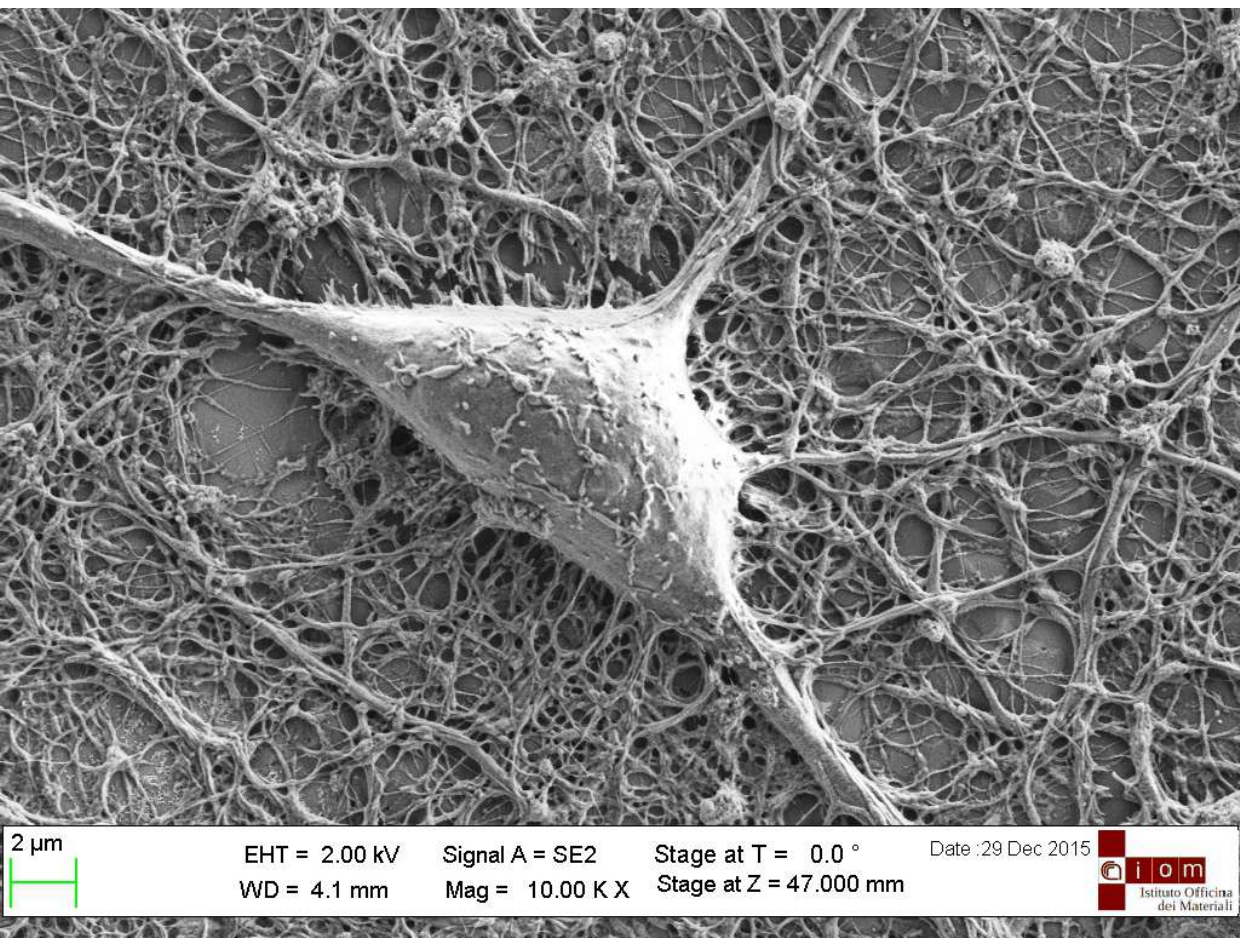}
     \includegraphics[width=0.22\textwidth]{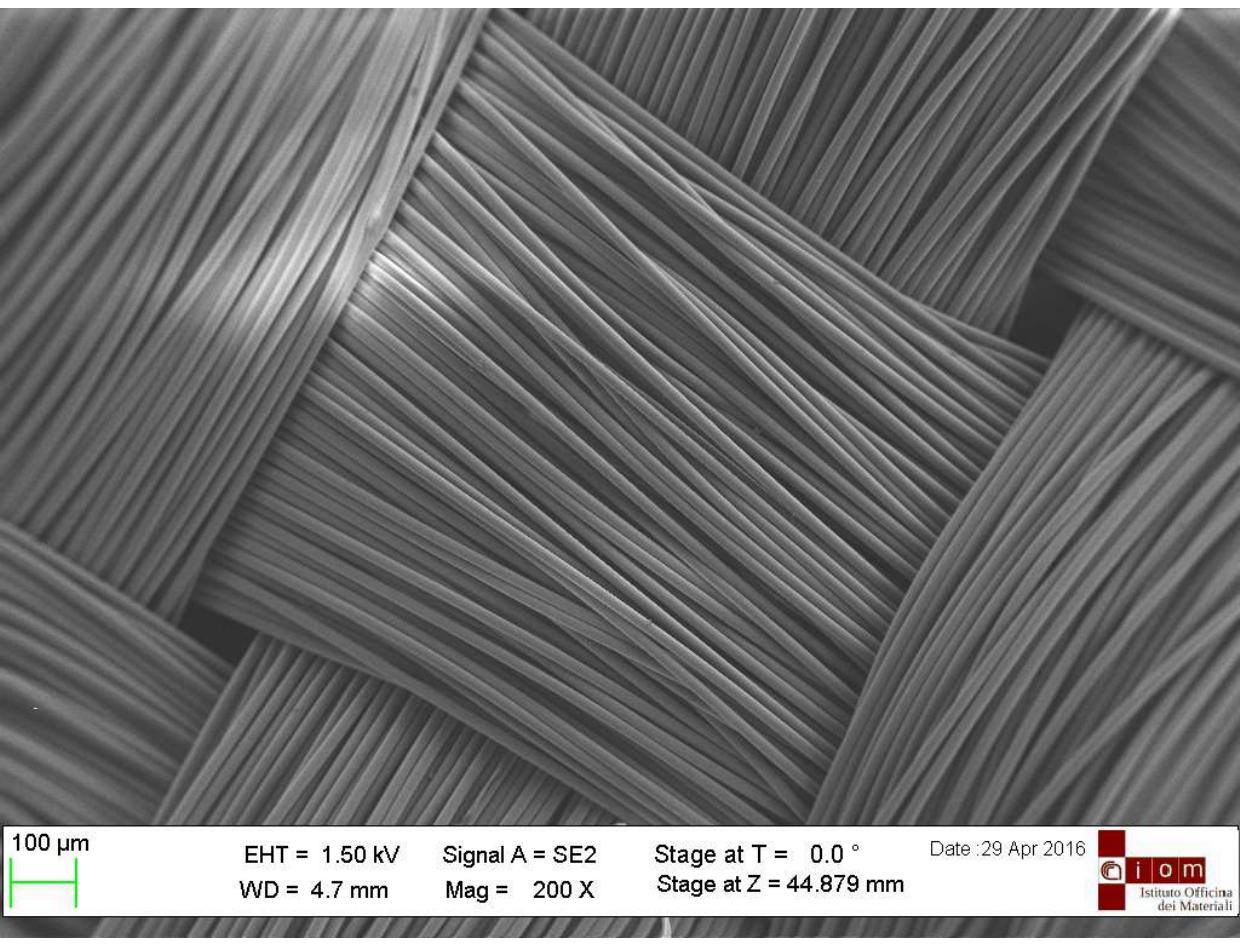}
     \includegraphics[width=0.22\textwidth]{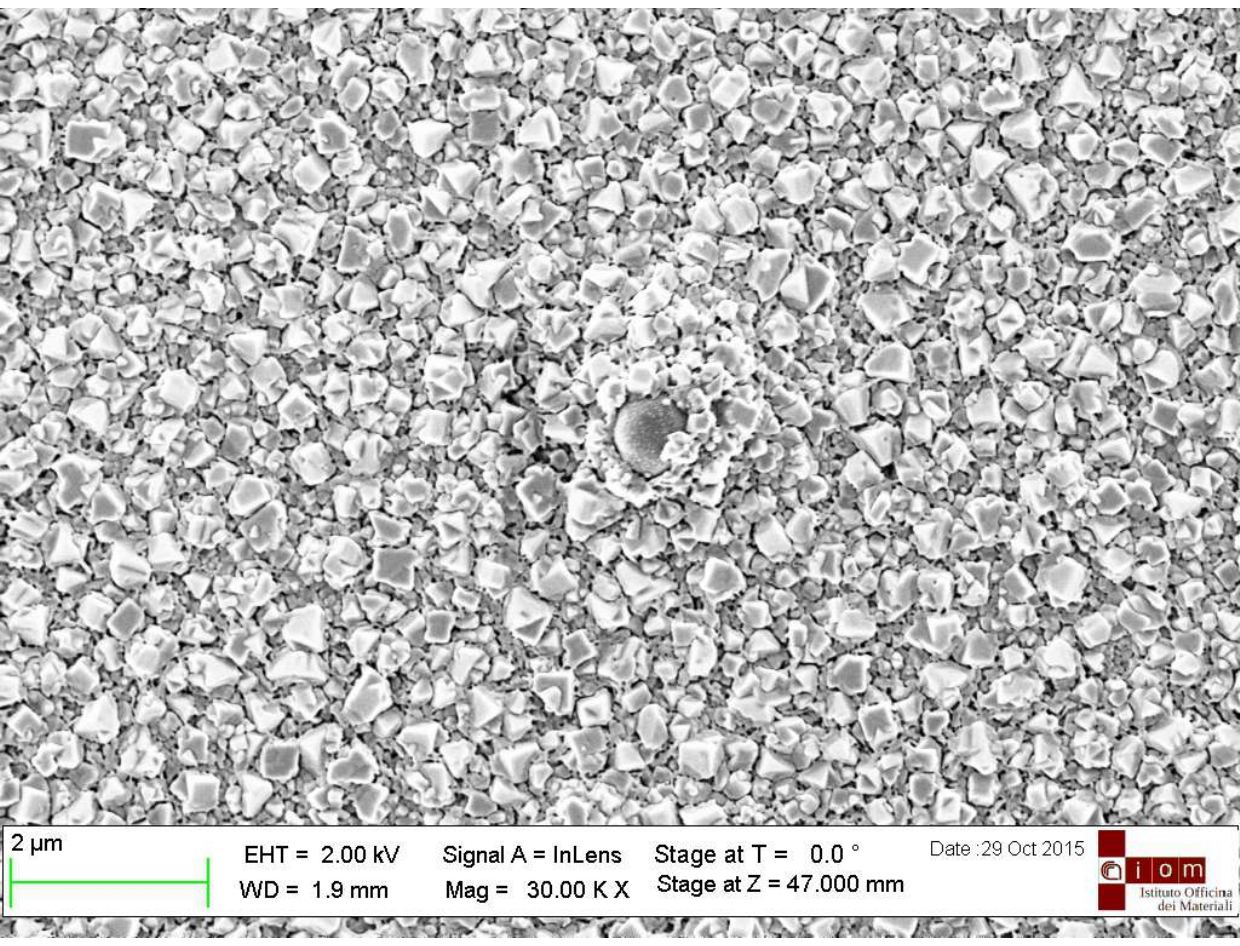}
     \includegraphics[width=0.22\textwidth]{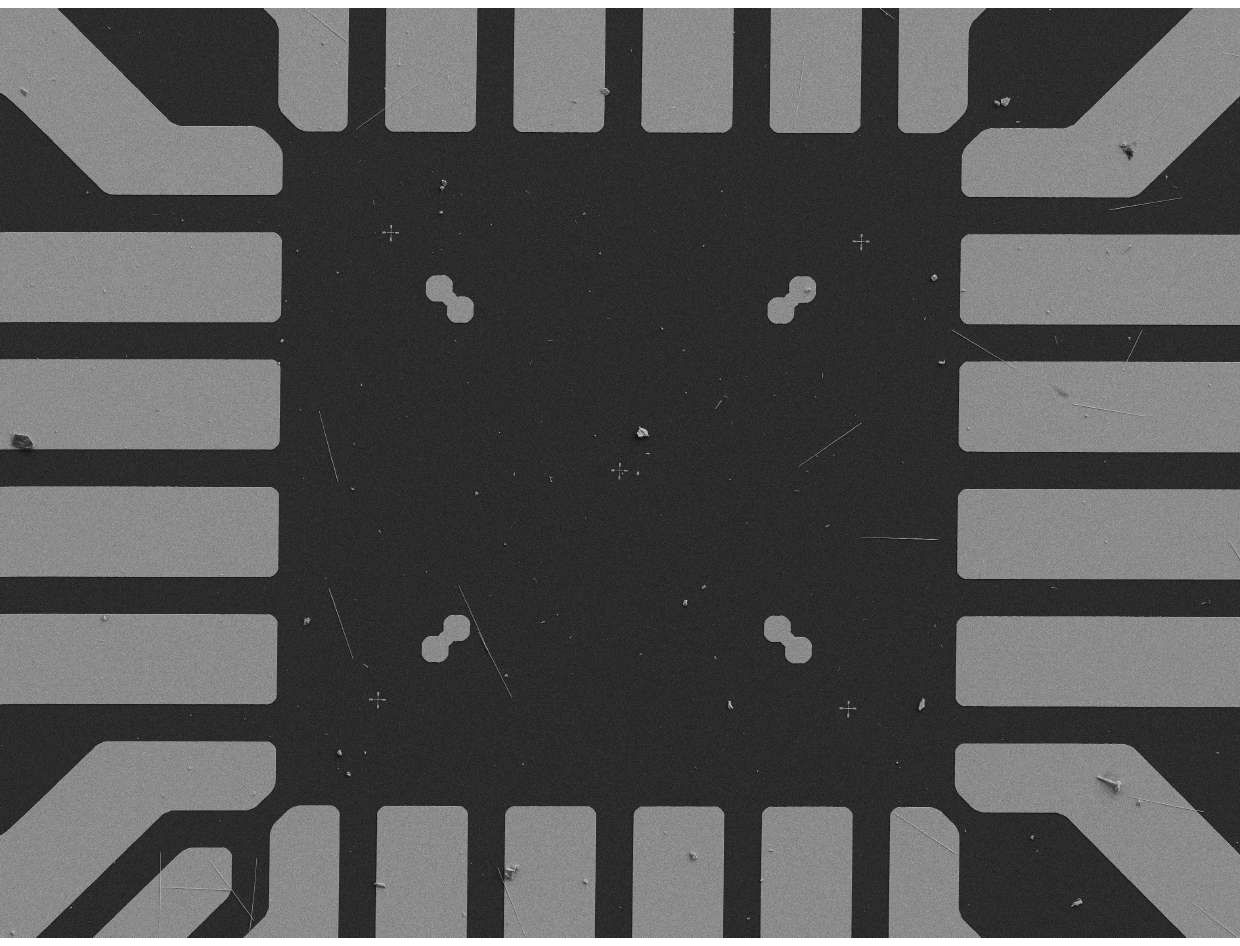}
     \includegraphics[width=0.22\textwidth]{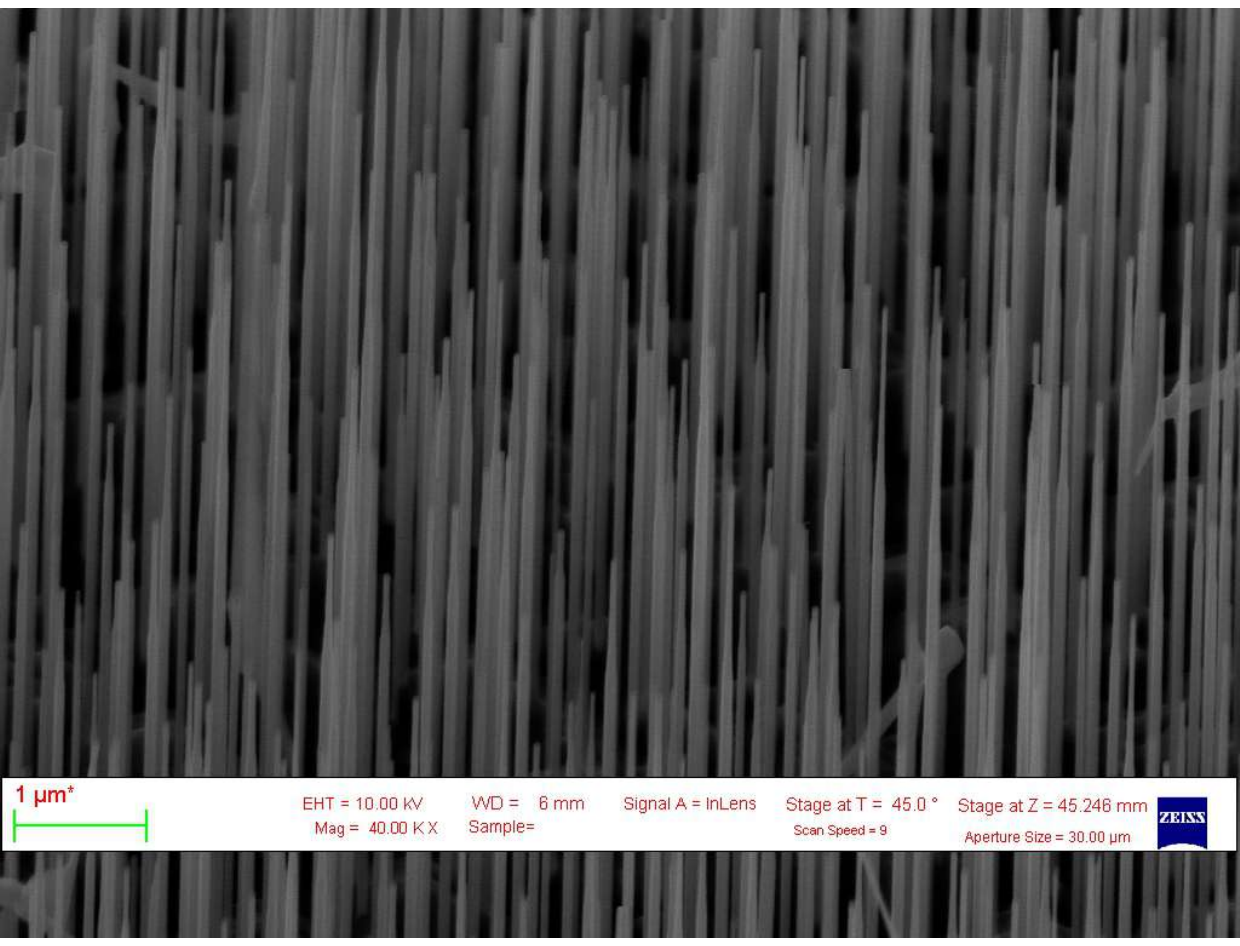}
     }
     \vspace{-15mm}
     \qquad
     \subfloat{\hspace{-10mm}\includegraphics[width=0.22\textwidth]{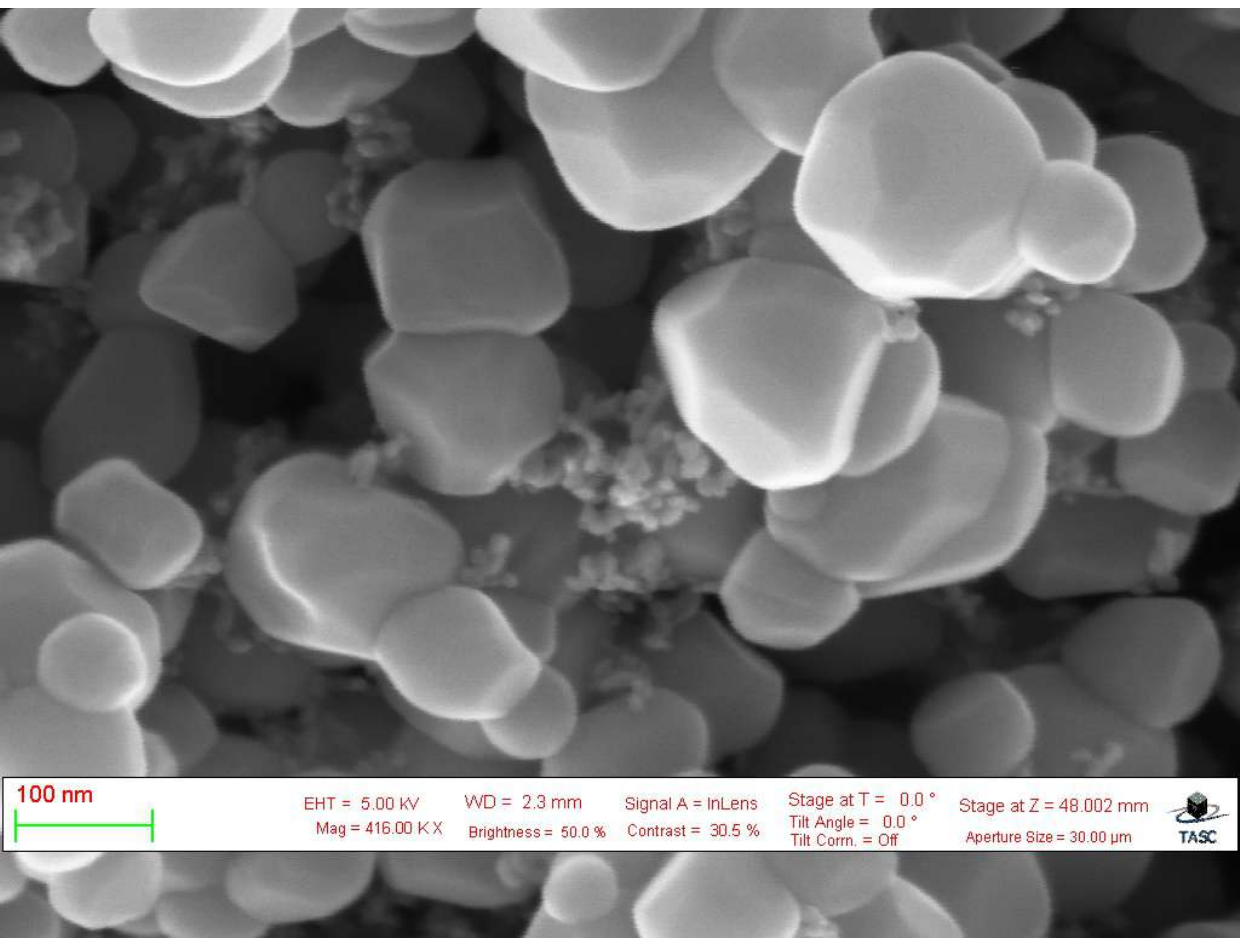}
     \includegraphics[width=0.22\textwidth]{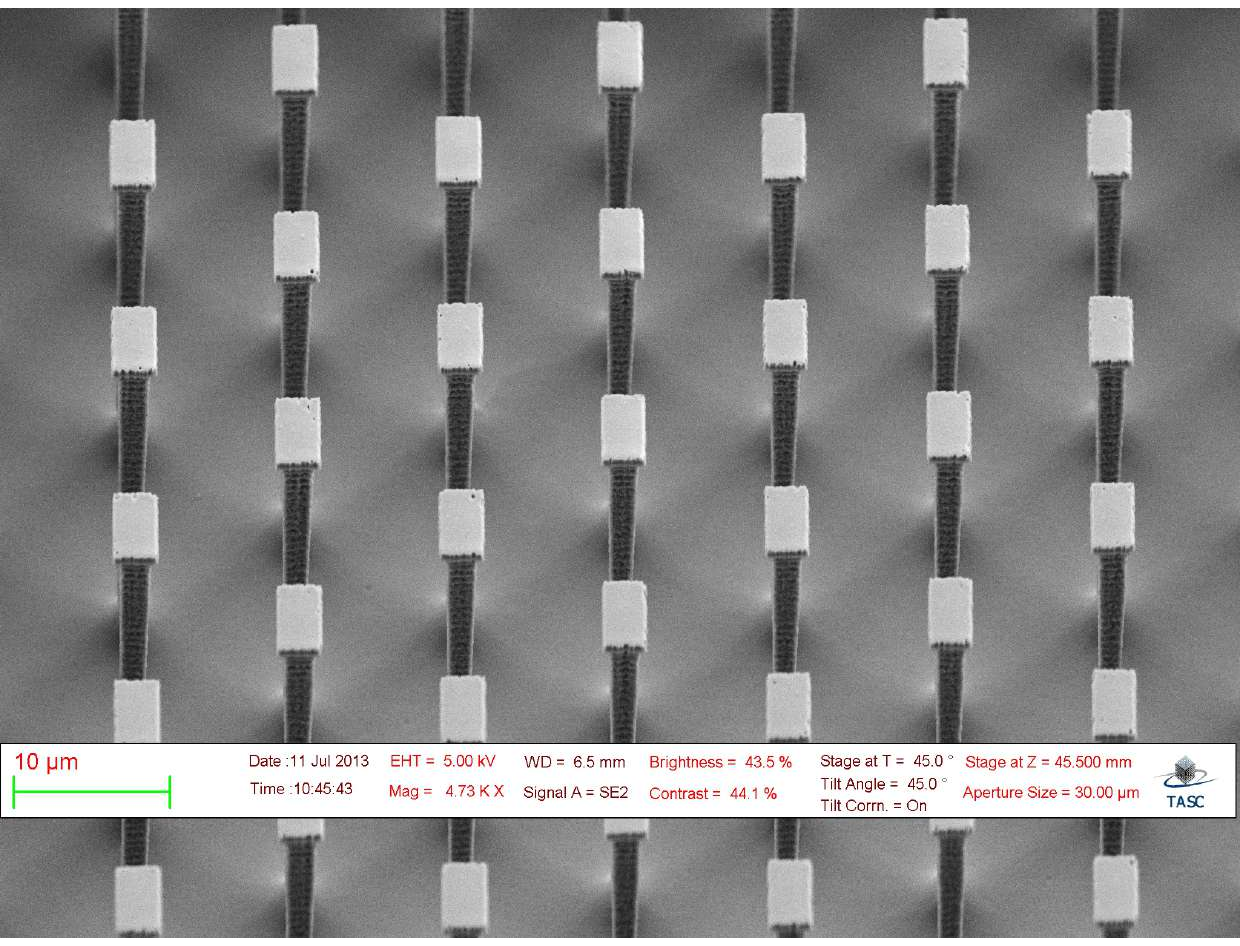}
     \includegraphics[width=0.22\textwidth]{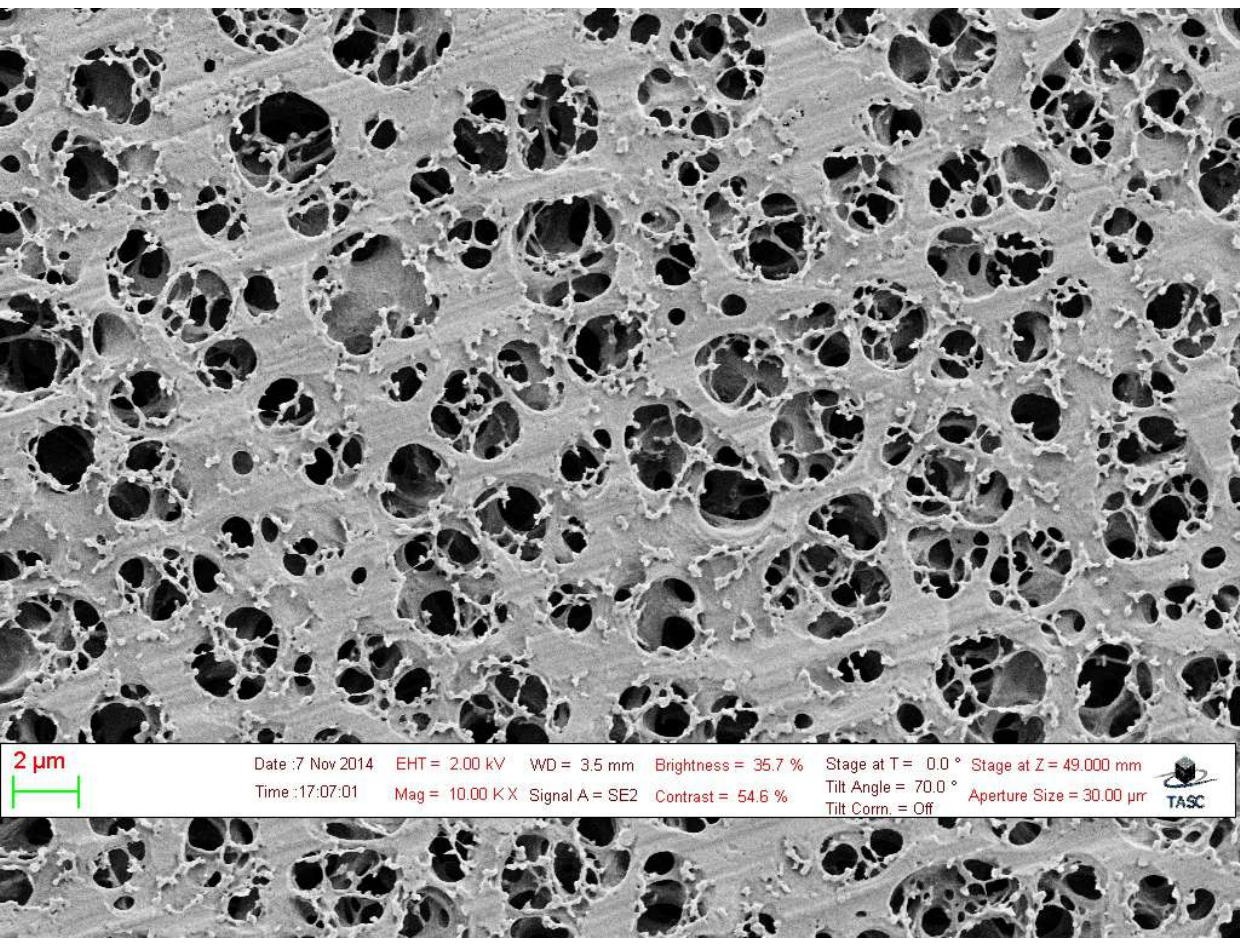}
     \includegraphics[width=0.22\textwidth]{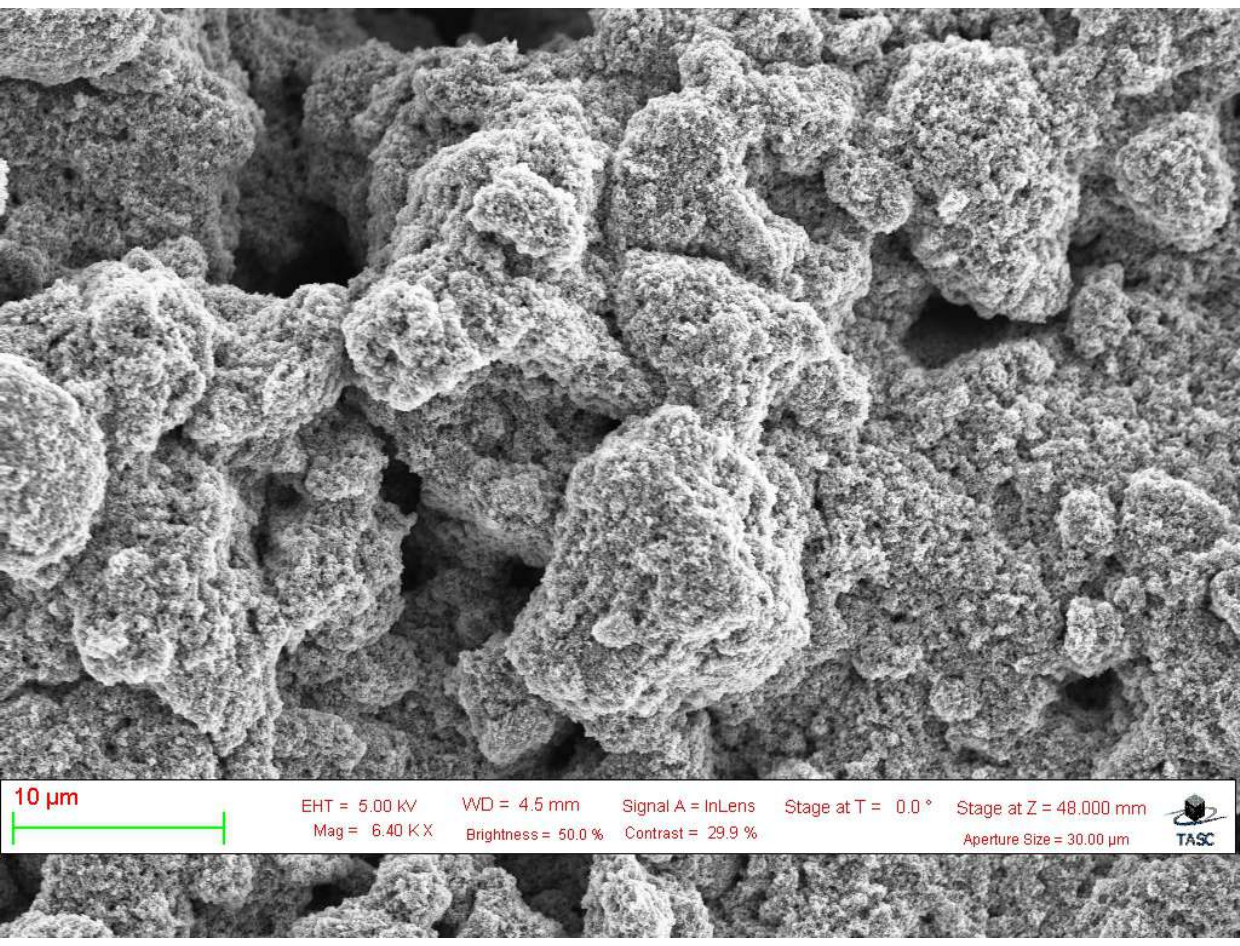}
     \includegraphics[width=0.22\textwidth]{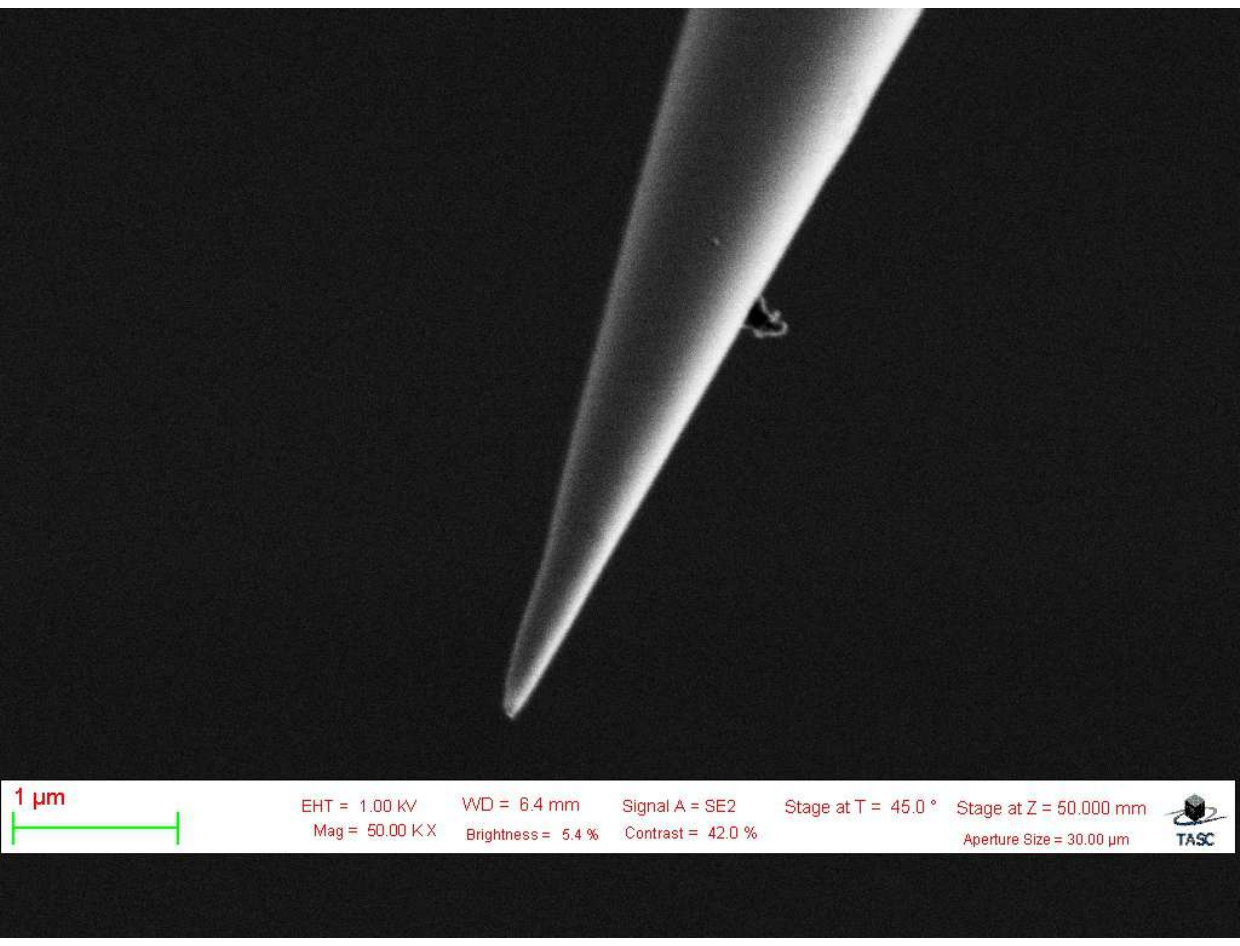}
     }
     \vspace{-10mm}
     \caption{The figure shows the electron micrograph categories in the SEM dataset (\cite{aversa2018first})(left to right in the first row: \textit{biological, fibers, films, MEMS, nanowires}; left to right in the second row: \textit{particles, patterned surface, porous sponges, powder, tips}).}
      \vspace{1mm}
     \label{fig:illustrationpics}
\end{figure}

\vspace{-1mm}
The SEM dataset contains ten diverse electron micrograph categories of nanomaterials such as \textit{biological, fibers, films, MEMS, patterned surfaces, etc}. Table ~\ref{tab:SEM} shows the unequal distributions in the total count of each electron micrograph category.  A few illustrative electron micrographs belonging to the different nanomaterials had shown in Figure \ref{fig:illustrationpics}.

\vspace{-4mm}

\begin{table}[htbp]
\footnotesize
\centering
\resizebox{0.525\textwidth}{!}{%
\subfloat{%
\begin{tabular}{|c|c|}
\hline
\textbf{Category}           & \textbf{Number of images} \\ \hline
Biological                  & 973($\text{4.57\%}$)                       \\
Tips                        & 1625($\text{7.63\%}$)                      \\
Fibres                      & 163($\text{0.76\%}$)                       \\
Porous Sponge               & 182($\text{0.85\%}$)                       \\
Films Coated Surface        & 327($\text{1.53\%}$)                       \\
Patterned surface           & 4756($\text{22.34\%}$)                      \\
Nanowires                   & 3821($\text{17.95\%}$)                      \\
Particles                   & 3926($\text{18.44\%}$)                      \\
MEMS devices and electrodes & 4591($\text{21.57\%}$)                      \\
Powder                      & 918($\text{4.31\%}$)                       \\ \hline
Total                       & 21282                     \\ \hline
\end{tabular}}}
\caption{Summary of SEM dataset: The listing of electron micrographs count per category.}
\label{tab:SEM}
\end{table}

\vspace{-5mm}

\vspace{-2mm}

\newpage

\vspace{-5mm}

\subsubsection{Benchmarking on Open-Source Material datasets}
\begin{figure}[htbp]
    \centering
    \includegraphics[scale=.45]{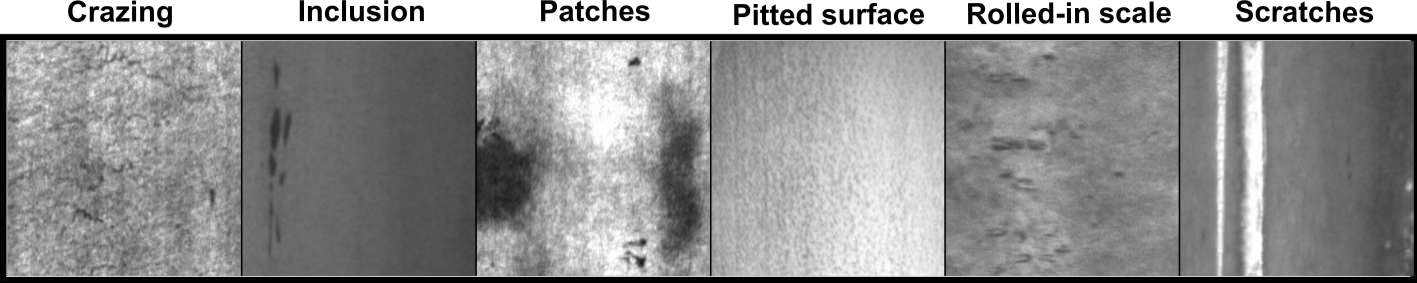}
    \vspace{-1mm}
    \caption{The NEU-SDD dataset: We illustrate the electron micrographs of six defect categories of hot-rolled steel strip \ref{note1}(\cite{deshpande2020one}).}
    \label{fig:NUE}
\end{figure}

\begin{figure}[htbp]
    \centering
    \includegraphics[scale=0.6]{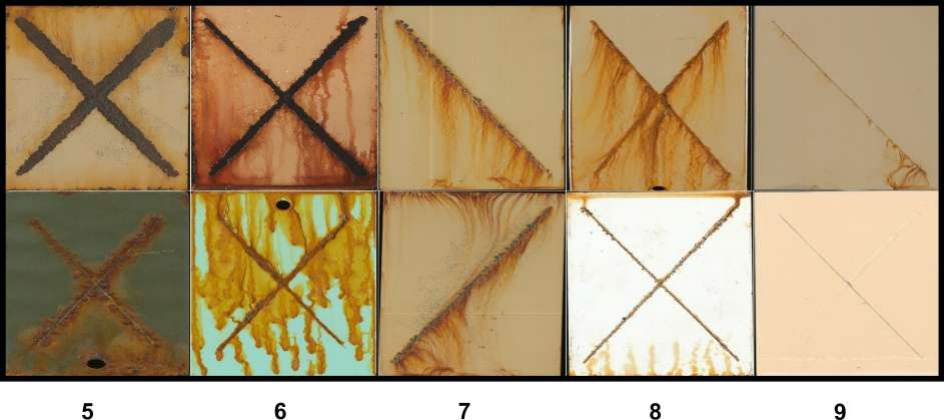}
    \vspace{-1mm}
    \caption{The CMI dataset: The representative electron  micrographs of five corrosion rating classes \ref{note2}.}
    \label{fig:corrosion}
\end{figure}

\begin{figure}[htbp]
    \centering
    \includegraphics[scale=0.6]{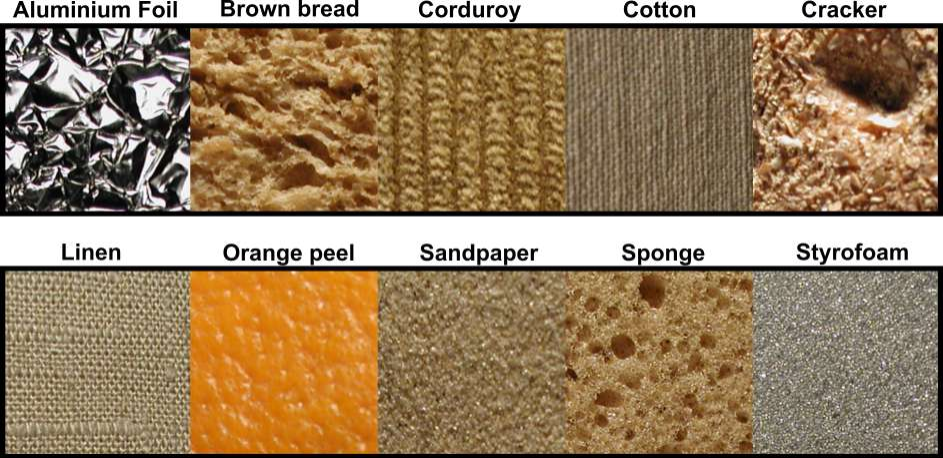}
    \vspace{-1mm}
    \caption{The KTH-TIPS dataset: Illustration of electron micrographs belonging to ten different materials \ref{note3}.}
    \label{fig:KTH}
\end{figure}

\begin{itemize}
    \item \textbf{NEU-SDD\footnote{Datasource: \url{http://faculty.neu.edu.cn/yunhyan/NEU_surface_defect_database.html}\label{note1}}
    }(\cite{deshpande2020one}) is a hot-rolled steel-strip surface defects database consisting of 1800 gray-scale electron micrographs with a resolution of 200$\times$200 pixels. The dataset had categorized into six surface defect classes, where each class has 300 electron micrographs. The list includes pitted surfaces, scratches, rolled-in scale, crazing, patches, and inclusion defects.  For illustration, each category’s representative electron micrograph had shown in Figure \ref{fig:NUE}. We compare the performance of our proposed method with several baseline algorithms on the multiclass classification task.    
    \item \textbf{CMI\footnote{\url{https://arl.wpi.edu/corrosion_dataset}\label{note2}}
    } is a benchmark dataset for automating corrosion assessment of materials, which consists of 600 corroding panel-based electron micrographs of resolution 512$\times$512 pixels. The human-annotated dataset had manually assigned a rating to each electron micrograph by the corrosion experts following the ASTM-D1654 standards. The discrete rating variables range from 5 to 9, where the corrosion rating of 9 implies that the panel is in the initial stage of corrosion. Each corrosion rating-based panel has 120 electron micrographs, and Figure \ref{fig:corrosion} shows the representative electron micrographs for each corrosion rating. We report the performance of our proposed method compared to the several baseline algorithms on the multiclass classification task.    
    \item \textbf{KTH-TIPS\footnote{\url{https://www.csc.kth.se/cvap/databases/kth-tips/index.html}\label{note3}}
    } is a texture database that contains 810 electron micrographs of ten different materials. The electron micrograph resolution in the dataset is 200$\times$200 pixels. In each material category, the corresponding electron micrographs have varying illumination, pose, and scale of the material under examination. The ten materials categories are \textit{sponge, orange peel, styrofoam, cotton, cracker, linen, brown bread, sandpaper, crumpled aluminum foil, and corduroy}. A few example electron micrographs per category had shown in Figure \ref{fig:KTH}. We evaluate and report the performance of our proposed method compared to the several baseline algorithms on the multiclass classification task.    
\end{itemize}
\vspace{-2mm}
Table \ref{table:P1} shows the performance comparison of our proposed method w.r.t to the baselines across all the datasets. The experimental results demonstrate the efficacy of our proposed method.

\vspace{-4mm}
\begin{table}[htbp]
\centering
\resizebox{0.5\textwidth}{!}{%
\subfloat{%
\setlength{\tabcolsep}{3pt}
\begin{tabular}{cc|cccc}
\hline
\multicolumn{2}{c|}{\textbf{Algorithms}}                                       & \textbf{NEU-SDD} & \textbf{CMI} & \textbf{KTH-TIPS}  \\ \hline
\multicolumn{1}{c|}{\multirow{4}{*}{\rotatebox[origin=c]{90}{\textbf{Baselines}}}} & ResNet                   & 0.906	& 0.928	& 0.941 &             \\
\multicolumn{1}{c|}{}                                          & GoogleNet                & 0.936	& 0.928	& 0.929
              \\
\multicolumn{1}{c|}{}                                          & SqueezeNet                & 0.955	& 0.943	& 0.963
              \\ 
\multicolumn{1}{c|}{}                                          & VanillaViT               & 0.962	& 0.968	& 0.972
 \\ 
\hline
\multicolumn{1}{c|}{}                                          & \textbf{Vision-HGNN}                  &     \textbf{0.975}              &      \textbf{0.981}             &    \textbf{0.987}               &                     \\ \hline
\end{tabular}}}
\vspace{-1mm}
\caption{Performance comparison on the datasets.}
\label{table:P1}
\end{table}

\vspace{-4mm}
\vspace{-4mm}
\vspace{-4mm}

\end{document}